\documentclass{article} 

\usepackage{microtype}
\usepackage{graphicx}
\usepackage{subfigure}
\usepackage{booktabs} 

\usepackage[accepted]{icml2025}

\usepackage[utf8]{inputenc} 
\usepackage[T1]{fontenc}    
\usepackage{hyperref}       
\usepackage{url}            
\usepackage{booktabs}       
\usepackage{amsfonts}       
\usepackage{nicefrac}       
\usepackage{microtype}      
\usepackage{xcolor}         
\usepackage{amsmath}
\usepackage{yfonts}
\usepackage{stmaryrd}
\usepackage{graphicx}
\usepackage{wrapfig}
\usepackage{algpseudocode}
\usepackage{mathtools}
\usepackage{amssymb}
\usepackage{multirow}
\usepackage{tikz}
\usetikzlibrary{arrows.meta}
\newtheorem{theorem}{Theorem}
\newtheorem{lemma}{Lemma}

\title{Inverse Flow and Consistency Models}

\begin{document}

\twocolumn[
\icmltitle{Inverse Flow and Consistency Models}




\begin{icmlauthorlist}
\icmlauthor{Yuchen Zhang}{utsw}
\icmlauthor{Jian Zhou}{utsw}
\end{icmlauthorlist}

\icmlaffiliation{utsw}{Lyda Hill Department of Bioinformatics, University of Texas Southwestern Medical Center, USA.}

\icmlcorrespondingauthor{Jian Zhou}{jian.zhou@utsouthwestern.edu}

\icmlkeywords{consistency models, flow matching, diffusion models, inverse problem}

\vskip 0.3in
]



\printAffiliationsAndNotice{}  

\begin{abstract}
Inverse generation problems, such as denoising without ground truth observations, is a critical challenge in many scientific inquiries and real-world applications. While recent advances in generative models like diffusion models, conditional flow matching, and consistency models achieved impressive results by casting generation as denoising problems, they cannot be directly used for inverse generation without access to clean data. Here we introduce Inverse Flow (IF), a novel framework that enables using these generative models for inverse generation problems including denoising without ground truth. Inverse Flow can be flexibly applied to nearly any continuous noise distribution and allows complex dependencies. We propose two algorithms for learning Inverse Flows, Inverse Flow Matching (IFM) and Inverse Consistency Model (ICM). Notably, to derive the computationally efficient, simulation-free inverse consistency model objective, we generalized consistency training to any forward diffusion processes or conditional flows, which have applications beyond denoising. We demonstrate the effectiveness of IF on synthetic and real datasets, outperforming prior approaches while enabling noise distributions that previous methods cannot support. Finally, we showcase applications of our techniques to fluorescence microscopy and single-cell genomics data, highlighting IF's utility in scientific problems. Overall, this work expands the applications of powerful generative models to inversion generation problems.

\end{abstract}

\section{Introduction}

\begin{figure*}[!h]
    \centering
    \includegraphics[width=1.0\textwidth]{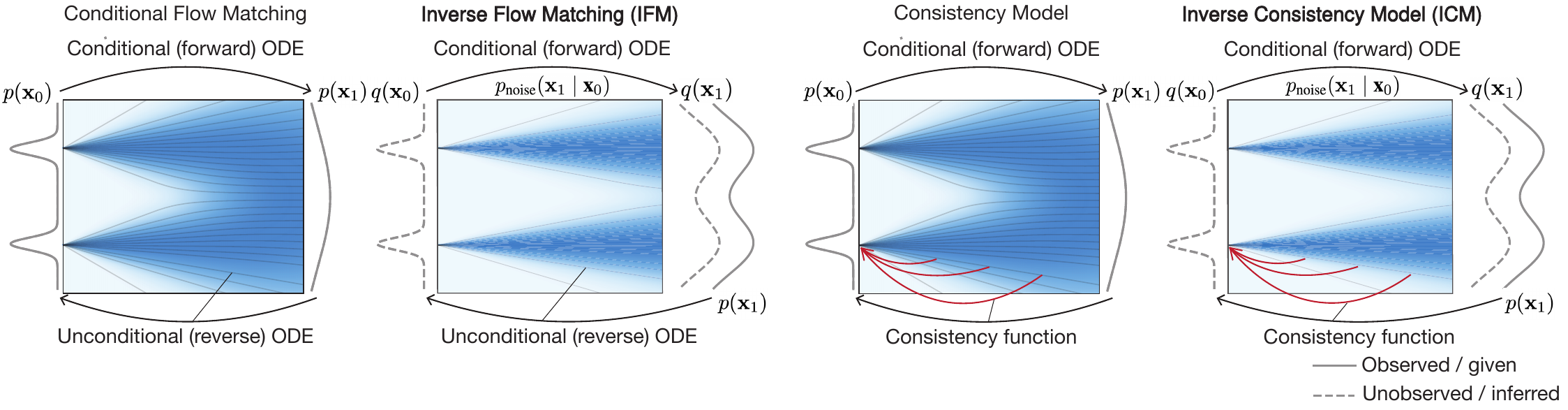}
  \vskip -0.2in

    \caption{Inverse flow enables adapting the family continuous-time generative models for solving inverse generation problems. For inverse flow matching and inverse consistency model, $\mathbf{x}_0$ indicates unobserved data and $\mathbf{x}_1$ indicates observed data. For conditional flow matching and consistency model,  $\mathbf{x}_0$ indicates data and $\mathbf{x}_1$ indicates variable from the prior distribution. Inverse flow algorithms modify continuous-time generative models to solve the inverse generation problem of recovering unobserved $\mathbf{x}_0$ from $\mathbf{x}_1$ by replacing the unobserved $p(\mathbf{x}_0)$ with generated $q(\mathbf{x}_0)$ within the training loop.}
    \label{fig:schema}
    \vskip -0.2in

\end{figure*}

Recent advances in generative modeling such as diffusion models \citep{sohl-dicksteinDeepUnsupervisedLearning2015,hoDenoisingDiffusionProbabilistic2020,songGenerativeModelingEstimating2020,songScoreBasedGenerativeModeling2021b,songDenoisingDiffusionImplicit2022}, conditional flow matching models \citep{lipmanFlowMatchingGenerative2023,tongImprovingGeneralizingFlowbased2024}, and consistency models \citep{songConsistencyModels2023,songImprovedTechniquesTraining2023} have achieved great success by learning a mapping from a simple prior distribution to the data distribution through an Ordinary Differential Equation (ODE) or Stochastic Differential Equation (SDE). We refer to their models as continuous-time generative models. These models typically involve defining a forward process, which transforms the data distribution to the prior distribution over time, and generation is achieved through learning a reverse process that can gradually transform the prior distribution to the data distribution (Figure \ref{fig:schema}).

Despite that those generative models are powerful tools for modeling the data distribution, they are not suitable for the \textit{inverse generation problems} when the data distribution is not observed and only data transformed by a forward process is given, which is typically true for noisy real-world data measurements. Mapping from noisy data to the latent ground truth is especially important in various scientific applications when pushing the limit of measurement capabilities. This limitation necessitates the exploration of novel methodologies that can bridge the gap between generative modeling and effective denoising in the absence of clean data.

Here we propose a new approach called \textbf{Inverse Flow (IF)}\footnote{Code available at \url{https://github.com/jzhoulab/InverseFlow}}, that learns a mapping from the observed noisy data distribution to the unobserved, ground truth data distribution (Figure \ref{fig:schema}), \textbf{inverting} the data requirement of generative models. An ODE or SDE is specified to reflect knowledge about the noise distribution. We further devised a pair of algorithms, \textbf{Inverse Flow Matching (IFM)} and \textbf{Inverse Consistency Model (ICM)} for learning inverse flows. Specifically, ICM involves a computationally efficient simulation-free objective that does not involve any ODE solver. 

A main contribution of our approach is generalizing continuous-time generative models to inverse generation problems such as denoising without ground truth. In addition, in order to develop ICM, we generalized the consistency training objective for consistency models to any forward diffusion process or conditional flow. This broadens the scope of consistency model applications and has implications beyond denoising. 

Compared to prior approaches for denoising without ground truth,  IF offers the most flexibility in noise distribution, allowing almost any continuous noise distributions including those with complex dependency and transformations. IF can be seamlessly integrated with generative modeling to generate samples from the ground truth rather than the observed noisy distribution. More generally, IF models the past states of a (stochastic) dynamical system before the observed time points using the knowledge of its dynamics, which can have applications beyond denoising.

\section{Background}

\subsection{Continuous-time generative models}

Our proposed inverse flow framework is built upon continuous-time generative models such as diffusion models, conditional flow matching, and consistency models. Here we present a unified view of these methods that will help connect inverse flow with this entire family of models (Section \ref{section:theory}).

These generative modeling methods are connected by their equivalence to continuous normalizing flow or neural ODE \citep{chenNeuralOrdinaryDifferential2019}. They can all be considered as explicitly or implicitly learning the ODE that transforms between the prior distribution $p(\mathbf{x}_1)$ and the data distribution $p(\mathbf{x}_0)$
\begin{equation}\label{eq:probability_flow_ode}
\begin{aligned}
\mathrm{d} \mathbf{x}= \mathbf{u}_t(\mathbf{x})  \mathrm{d} t.
\end{aligned}
\end{equation}
in which  $\mathbf{u}_t(\mathbf{x})$ represents the vector field of the ODE. We use the convention that $t=0$ corresponds to the data distribution and $t=1$ corresponds to the prior distribution. Generation is realized by reversing this ODE, which makes this family of methods a natural candidate for extension toward denoising problems.

Continuous-time generative models typically involve defining a conditional ODE or SDE that determines the $p(\mathbf{x}_t| \mathbf{x}_0)$ that transforms the data distribution to the prior distribution. Training these models involves learning the unconditional ODE (Eq. \ref{eq:probability_flow_ode}) based on $\mathbf{x}_0$ and the conditional ODE or SDE \citep{lipmanFlowMatchingGenerative2023,tongImprovingGeneralizingFlowbased2024,songScoreBasedGenerativeModeling2021b} (Figure \ref{fig:schema}).  The unconditional ODE can be used for generation from noise to data.

\subsubsection{Conditional flow matching}\label{section:cfm}
Conditional flow matching defines the transformation from data to prior distribution via a conditional ODE vector field $\mathbf{u}_t(\mathbf{x}\mid \mathbf{x}_0)$. The unconditional ODE vector field  $\mathbf{v}^\theta_t(\mathbf{x})$ is learned by minimizing the objective \citep{lipmanFlowMatchingGenerative2023,tongImprovingGeneralizingFlowbased2024,albergoBuildingNormalizingFlows2023}:

\begin{equation}\label{eq:L_CFM_v1}
    \begin{aligned}
    \left\|\mathbf{v}^\theta_t(\mathbf{x}_{t})-\mathbf{u}_t\left(\mathbf{x}_{t} \mid \mathbf{x}_0\right)\right\|.
        \end{aligned}
\end{equation}
where $\mathbf{x}_0$ is sampled from the data distribution, and $\mathbf{x}_t$ is sampled from the conditional distribution $p(\mathbf{x}_t \mid \mathbf{x}_0)$ given by the conditional ODE. 

The conditional ODE vector field $\mathbf{u}_t(\mathbf{x}\mid \mathbf{x}_0)$ can also be stochastically approximated through sampling from both prior distribution and data distribution and using the conditional vector field $\mathbf{u}_t(\mathbf{x}\mid \mathbf{x}_0, \mathbf{x}_1)$ as the training target \citep{lipmanFlowMatchingGenerative2023,tongImprovingGeneralizingFlowbased2024}:

\begin{equation}\label{eq:L_CFM_v2}
    \begin{aligned}
    \left\|\mathbf{v}^\theta_t(\mathbf{x}_{t})-\mathbf{u}_t\left(\mathbf{x}_{t} \mid \mathbf{x}_0, \mathbf{x}_1\right)\right\|.
        \end{aligned}
\end{equation}

This formulation has the benefit that $\mathbf{u}_t(\mathbf{x}\mid \mathbf{x}_0, \mathbf{x}_1)$ can be easily chosen as any interpolation between $\mathbf{x}_0$ and $\mathbf{x}_1$, because this interpolation does not affect the probability density at time 0 or 1 \citep{lipmanFlowMatchingGenerative2023,tongImprovingGeneralizingFlowbased2024,albergoBuildingNormalizingFlows2023,albergoStochasticInterpolantsUnifying2023}. For example, a linear interpolation corresponds to $\mathbf{x}_t = \mathbf{x}_0 + t(\mathbf{x}_1-\mathbf{x}_0)$ \citep{lipmanFlowMatchingGenerative2023,tongImprovingGeneralizingFlowbased2024,liuFlowStraightFast2022}. Sampling is realized by simulating the unconditional ODE with learned vector field $\mathbf{v}^\theta_t(\mathbf{x})$ in the reverse direction.

\subsubsection{Consistency models}\label{section:consistency_models}

In contrast, consistency models \citep{songConsistencyModels2023,songImprovedTechniquesTraining2023} learn consistency functions that can directly map a sample from the prior distribution to data distribution,  equivalent to simulating the unconditional ODE in the reverse direction:
$$\mathbf{c}(\mathbf{x}_t, t) = \textrm{ODE}^\mathbf{u}_{t\rightarrow 0}(\mathbf{x}_t)$$
where $\mathbf{x}_t$ denotes $\mathbf{x}$ at time $t$, and we use $\textrm{ODE}^\mathbf{u}_{t\rightarrow 0}(\mathbf{x}_t)$ to denote simulating the ODE with vector field $\mathbf{u}_t(\mathbf{x})$ from time $t$ to time $0$ starting from $\mathbf{x}_t$. The consistency function is trained by minimizing the consistency loss \citep{songConsistencyModels2023}, which measures the difference between consistency function evaluations at two adjacent time points
\begin{equation}\label{eq:L_CM}
    \begin{aligned}
    &\mathcal{L}_{\mathrm{CM}}(\theta)=\\
    &\quad\mathbb{E}_{i, \mathbf{x}_{t_i}, \mathbf{x}_{t_{i+1}}}\left[ \left\|\mathbf{c}_\theta(\mathbf{x}_{t_{i+1}},t_{i+1})-\text{stopgrad}\left(\mathbf{c}_\theta (\mathbf{x}_{t_{i}},t_{i})\right) \right\|\right]
    \end{aligned}
\end{equation}
with the boundary condition $\mathbf{c}(\mathbf{x},0)=\mathbf{x}$. Stopgrad indicates that the term within the operator does not get optimized.

There are two approaches to training consistency models: one is distillation, and the other is training from scratch. In the consistency distillation objective, a pretrained diffusion model is used to obtain the unconditional ODE vector field $\mathbf{u}_{t}$, and $\mathbf{x}_{t_{i+1}}$ and $\mathbf{x}_{t_i}$ differs by one ODE step
\begin{equation}\label{eq:CD_sample}
    \begin{aligned}
        \mathbf{x}_{t_{i+1}} &\sim p(\mathbf{x}_{t_{i+1}} \mid \mathbf{x}_0),\\  \mathbf{x}_{t_{i+1}}-\mathbf{x}_{t_{i}} &= \mathbf{u}_{t_{i+1}}(\mathbf{x}_{t_{i+1}})(t_{i+1}-t_i)
    \end{aligned}
\end{equation}
If the consistency model is trained from scratch, the consistency training objective samples $\mathbf{x}_{t_{i+1}}$ and $\mathbf{x}_{t_{i}}$ in a coupled manner from the forward diffusion process \citep{karrasElucidatingDesignSpace2022b}
\begin{equation}\label{eq:Karras_sample}
    \begin{aligned}
    \mathbf{x}_{t_{i+1}} = \mathbf{x}_0 + \mathbf{z} {t_{i+1}}, \quad \mathbf{x}_{t_{i}} = \mathbf{x}_0 + \mathbf{z} {t_{i}} ,\quad \mathbf{z}\sim \mathcal{N}(0, \sigma^2 \mathbf{I})
    \end{aligned}
\end{equation}
where $\sigma$ controls the maximum noise level at $t=1$. Consistency models have the advantage of fast generation speed as they can generate samples without solving any ODE or SDE.

\subsubsection{Diffusion models}\label{section:diffusion_models}
In diffusion models, the transformation from data to prior distribution is defined by a forward diffusion process (conditional SDE). The diffusion model training learns the score function which determines the unconditional ODE, also known as the probability flow ODE \citep{songScoreBasedGenerativeModeling2021b}.
\paragraph{Denoising applications of diffusion models}
Diffusion models are inherently connected to denoising problems as the generation process is essentially a denoising process. However, existing denoising methods using diffusion models require training on ground truth data \citep{yue_resshift_2023,xie_diffusion_2023}, which is not available in inverse generation problems. 
\paragraph{Ambient diffusion and GSURE-diffusion} 
Ambient Diffusion \citep{daras_ambient_2023} and GSURE-diffusion \citep{kawar_gsure-based_2024} address a related problem of learning the distribution of clean data by training on only linearly corrupted (linear transformation followed by additive Gaussian noise) data. Although those methods are designed for generation, they can be applied to denoising. 
Ambient Diffusion Posterior Sampling \citep{aali_ambient_2024}, further allowed using models trained with ambient diffusion on corrupted data to perform posterior sampling-based denoising for a different forward process (e.g., blurring). Consistent Diffusion Meets Tweedie \citep{daras_consistent_2024} improves Ambient Diffusion by allowing exact sampling from clean data distribution using consistency loss with a double application of Tweedie's formula. \cite{rozet_learning_2024} explored the potential of expectation maximization in training diffusion models on corrupted data. However, all these methods are restricted to training on linearly corrupted data, which still limit their applications when the available data is affected by other types of noises.

\subsection{Denoising without ground truth}

Denoising without access to ground truth data requires assumptions about the noise or the signal. Most contemporary approaches are based on assumptions about the noise, as the noise distribution is generally much simpler and better understood. Because prior methods have been comprehensively reviewed \citep{kimNoise2ScoreTweedieApproach2021,batsonNoise2SelfBlindDenoising2019,lehtinenNoise2NoiseLearningImage2018,xieNoise2SameOptimizingSelfSupervised2020,soltanayevTrainingDeepLearning2018,metzlerUnsupervisedLearningStein2020}, and our approach is not directly built upon these approaches, we only present a brief overview and refer the readers to Appendix \ref{appendix:denoise} referenced literature for more detailed discussion. None of these approaches are generally applicable to any noise types.

\section{Inverse Flow and Consistency Models}\label{section:theory}

In continuous-time generative models, usually the data $\mathbf{x}_0$ from the distribution of interest is given. In contrast, in inverse generation problems, only the transformed data $\mathbf{x}_1$ and the conditional distribution $p(\mathbf{x}_1|\mathbf{x}_0)$ are given, whereas $\mathbf{x}_0$ are unobserved. For example, $\mathbf{x}_1$ are the noisy observations and $p(\mathbf{x}_1|\mathbf{x}_0)$ is the conditional noise distribution. We define the \textit{Inverse Flow} (IF) problem as finding a mapping from $\mathbf{x}_1$ to $\mathbf{x}_0$ which allows not only recovering the unobserved data distribution $p(\mathbf{x}_0)$ but also providing an estimate of $\mathbf{x}_0$ from $\mathbf{x}_1$ (Figure \ref{fig:schema}).

For denoising without ground truth applications, the inverse flow framework requires only the noisy data $\mathbf{x}_1$ and the ability to sample from the noise distribution $p(\mathbf{x}_1|\mathbf{x}_0)$. This is thus applicable to any continuous noise and allows complex dependencies on the noise distribution, including noise that can only be sampled through a diffusion process.

Intuitively, without access to unobserved data $\mathbf{x}_0$, inverse flow algorithms train a continuous-time generative model using generated $\mathbf{x}_0$ from observed data $\mathbf{x}_1$ within the training loop (Figure \ref{fig:schema}). We demonstrated that this approach effectively recovers the unobserved distribution $p(\mathbf{x}_0)$ and learns a mapping from $\mathbf{x}_1$ to $\mathbf{x}_0$.

\subsection{Inverse Flow Matching}

To solve the inverse flow problem, 
we first consider learning a mapping from $\mathbf{x}_{1}$ to $\mathbf{x}_{0}$ through an ODE with vector field $\mathbf{v}^{\theta}_t(\mathbf{x})$. We propose to learn $\mathbf{v}^{\theta}_t(\mathbf{x})$ with the inverse flow matching (IFM) objective

\begin{equation}\label{eq:L_IFM_v1}
    \begin{aligned}
    &\mathcal{L}_{\mathrm{IFM}}(\theta)\\
    &\quad=\mathbb{E}\left\|\mathbf{v}^{\theta}_t(\mathbf{x}_{t})-\mathbf{u}_t\left(\mathbf{x}_{t} \mid \textrm{ODE}^{\mathbf{v}^{\theta}}_{1\rightarrow 0}(\mathbf{x}_1)\right)\right\|
        \end{aligned}
\end{equation}

where the expectation is taken over $t$, $p(\mathbf{x}_1)$, and $p(\mathbf{x}_{t} \mid \mathbf{x}_0=\textrm{ODE}^{\mathbf{v}^{\theta}}_{1\rightarrow 0}(\mathbf{x}_1))$. This objective differs from conditional flow matching (Eq. \ref{eq:L_CFM_v1}) in two key aspects: using only transformed data $\mathbf{x}_1$ rather than unobserved data $\mathbf{x}_0$, and choosing the conditional ODE based on the conditional distribution $p(\mathbf{x}_1|\mathbf{x}_0)$. Specifically,
\begin{enumerate}
    \item Sampling from the data distribution $p(\mathbf{x}_0)$ is replaced with sampling from $p(\mathbf{x}_1)$ and simulating the unconditional ODE backward in time based on the vector field $\mathbf{v}$, denoted as $\textrm{ODE}^{\mathbf{v}^{\theta}}_{t\rightarrow 0}(\mathbf{x}_1)$. We refer to this distribution as the recovered data distribution $q(\mathbf{x}_0)$.
    \item The conditional ODE vector field $\mathbf{u}_t\left(\mathbf{x} \mid \mathbf{x}_0\right)$ is chosen to match the given conditional distribution $p(\mathbf{x}_1|\mathbf{x}_0)$ at time $1$.
\end{enumerate}

For easier and more flexible application of IFM, similar to conditional flow matching (Eq. \ref{eq:L_CFM_v2}), an alternative form of the conditional ODE $\mathbf{u}_t\left(\mathbf{x} \mid \mathbf{x}_0, \mathbf{x}_1'\right)$ can be used instead of $\mathbf{u}_t\left(\mathbf{x} \mid \mathbf{x}_0\right)$. Since $\mathbf{x}_1'$ is sampled from the noise distribution  $p(\mathbf{x}_1|\mathbf{x}_0)$, the above condition is automatically satisfied. The conditional ODE vector field can be easily chosen as any smooth interpolation between $\mathbf{x}_0$ and $\mathbf{x}_1'$, such as  $\mathbf{u}_t\left(\mathbf{x} \mid \mathbf{x}_0, \mathbf{x}_1'\right) = \mathbf{x}_1' - \mathbf{x}_0$. We detailed the inverse flow matching training in Algorithm \ref{alg:IFM} with the alternative form in Appendix \ref{appendix:alternative}.

Next, we discuss the theoretical justifications of the IFM objective and the interpretation of the learned model. We show below that when the loss converges, the recovered data distribution  $q\left(\mathbf{x}_0\right)$ matches the ground truth distribution  $p(\mathbf{x}_0)$. The proof is provided in Appendix \ref{appendix:IFM}.

\begin{theorem}\label{theorem:IFM}
    Assume that the noise distribution $p(\mathbf{x}_{1}\mid \mathbf{x}_{0})$ satisfies the condition that,  for any noisy data distribution $p(\mathbf{x}_{1})$ there exists only one probability distribution $p(\mathbf{x}_{0})$ that satisfies $p(\mathbf{x}_{1})=\int p(\mathbf{x}_{1}\mid \mathbf{x}_{0})p(\mathbf{x}_{0})\mathrm{d}\mathbf{x}_0$ , then under the condition that $\mathcal{L}_{\mathrm{IFM}}=0$,  we have the recovered data distribution $q(\mathbf{x}_{0})=p(\mathbf{x}_{0})$. 
\end{theorem}

Moreover, we show that with IFM the learned ODE trajectory from $\mathbf{x}_1$ to $\mathbf{x}_0$ can be intuitively interpreted as always pointing toward the direction of the estimated $\mathbf{x}_0$. More formally, the learned unconditional ODE vector field can be interpreted as an expectation of the conditional ODE vector field. 
\begin{lemma}\label{lemma:1}
    Given a conditional ODE vector field $\mathbf{u}_{t}(\mathbf{x}\mid \mathbf{x}_{0}, \mathbf{x}_{1})$ that generates a conditional probability path $p(\mathbf{x}_{t}\mid \mathbf{x}_{0}, \mathbf{x}_{1})$, the unconditional probability path $p(\mathbf{x}_{t})$ can be generated by the unconditional ODE vector field $\mathbf{u}_{t}(\mathbf{x})$, which is defined as
\begin{equation}\label{eq:flowexpectation}
    \begin{aligned}
        \mathbf{u}_{t}(\mathbf{x})=\mathbb{E}_{p(\mathbf{x}_{0},\mathbf{x}_{1}\mid \mathbf{x})}\left[\mathbf{u}_{t}(\mathbf{x}\mid \mathbf{x}_{0},\mathbf{x}_{1})\right]
    \end{aligned}
\end{equation}
\end{lemma}
\vskip -0.12in
The proof is provided in Appendix \ref{appendix:IFM}. 
Specifically, with the choice of $\mathbf{u}_t\left(\mathbf{x} \mid \mathbf{x}_0, \mathbf{x}_1\right) = \mathbf{x}_1 - \mathbf{x}_0$, Eq. \ref{eq:flowexpectation} has an intuitively interpretable form
\begin{equation}
    \begin{aligned}
        \mathbf{u}_{t}(\mathbf{x})=\mathbb{E}_{p(\mathbf{x}_{0}\mid \mathbf{x})}\left[\frac{\mathbf{x} - \mathbf{x}_{0}}{t} \right]
    \end{aligned}
\end{equation}
\vskip -0.12in
which means that the unconditional ODE vector field at any time $t$ points straight toward the expected ground truth $\mathbf{x}_{0}$.

\subsection{Simulation-free Inverse Flow with Inverse Consistency Model}

IFM can be computationally expensive during training and inference because it requires solving ODE in each update. We address this limitation by introducing inverse consistency model (ICM), which learns a consistency function to directly solve the inverse flow without involving an ODE solver. 

However, the original consistency training formulation is specific to one type of diffusion process \citep{karrasElucidatingDesignSpace2022b},  which is \textit{only applicable to independent Gaussian noise distribution} for inverse generation application. Thus, to derive inverse consistency model that is applicable to any transformation, we first generalize consistency training so that it can be applied to arbitrary transformations and thus flexible to model almost any noise distribution.

\begin{figure*}
\begin{minipage}[t]{0.45\textwidth}
\begin{algorithm}[H]
    \centering
    \caption{IFM Training}\label{alg:IFM}
    \begin{algorithmic}[1]
        \State \textbf{Input:} dataset $\mathcal{D}$, initial model parameter $\theta$, and learning rate $\eta$
        \Repeat
        \State Sample $\mathbf{x}_{1}\sim \mathcal{D}$ and  $t\sim \mathcal{U}[0,1]$
        \State $\mathbf{x}_0 \leftarrow \textrm{stopgrad}\left(\textrm{ODE}^{\mathbf{v}^{\theta}}_{1\rightarrow 0}(\mathbf{x}_1)\right)$
        \State Sample $\mathbf{x}_{t}\sim p(\mathbf{x}_{t}\mid \mathbf{x}_{0})$
        \State $\mathcal{L}(\theta)\leftarrow\left\|\mathbf{v}^{\theta}_t(\mathbf{x}_{t})-\mathbf{u}_t\left(\mathbf{x}_{t} \mid \mathbf{x}_{0}\right)\right\|$
        \State  $\theta\leftarrow\theta-\eta\nabla_{\theta}\mathcal{L}(\theta)$
        \Until {convergence}
    \end{algorithmic}
\end{algorithm}
\end{minipage}
\begin{minipage}[t]{0.53\textwidth}
\begin{algorithm}[H]
    \centering
    \caption{ICM Training}\label{alg:ICM}
    \begin{algorithmic}[1]
        \State \textbf{Input:} dataset $\mathcal{D}$, initial model parameter $\theta$, learning rate $\eta$, and sequence of time points $0=t_{1}<t_{2}<\dots<t_{N}=1$
        \Repeat
        \State Sample $\mathbf{x}_{1}\sim \mathcal{D}$ and  $i\sim \mathcal{U}[1,N-1]$
        \State $\mathbf{x}_0 \leftarrow \textrm{stopgrad}\left(\mathbf{c}_{\theta}(\mathbf{x}_{1},1)\right)$
        \State Sample $\mathbf{x}_{t_{i+1}}\sim p(\mathbf{x}_{t_{i+1}}\mid \mathbf{x}_{0})$
        \State $\mathbf{x}_{t_{i}}\leftarrow\mathbf{x}_{t_{i+1}}-\mathbf{u}_{t_{i+1}}(\mathbf{x}_{t_{i+1}}\mid \mathbf{x}_0)(t_{i+1}-t_i)$
        \State $\begin{multlined}
            \mathcal{L}(\theta)\leftarrow
       \left\| \mathbf{c}_{\theta}(\mathbf{x}_{t_{i+1}},t_{i+1})-\textrm{stopgrad}\left(\mathbf{c}_{\theta}(\mathbf{x}_{t_{i}},t_{i}) \right)\right\|
        \end{multlined}
        $
        \State  $\theta\leftarrow\theta-\eta\nabla_{\theta}\mathcal{L}(\theta)$
        \Until {convergence}
    \end{algorithmic}
\end{algorithm}
\end{minipage}
\vskip -0.2in
\end{figure*}

\subsubsection{Generalized Consistency Training}\label{section:gct}

To recall from Section \ref{section:consistency_models}, consistency distillation is only applicable to distilling a pretrained diffusion or conditional flow matching model. The consistency training objective allows training consistency models from scratch but only for a specific forward diffusion process, which limits its flexibility in applying to any inverse generation problem.

\begin{center}
\begin{tikzpicture}
    \centering
    \node (A) at (0, 0.75) {\(\small{\textbf{Generalized Consistency Training}}\)};
    \node (B) at (4.5, 0.75) {\(\small{\textrm{Consistency Distillation}}\)};
    \node (C) at (0, -0.75) {\(\small{\textrm{Conditional Flow Matching}}\)};
    \node (D) at (4.5, -0.75) {\(\small{\textrm{Flow Matching}}\)};
    
    \draw[<->, thick] (A) -- (B);
    \draw[<->, thick] (C) -- (D);
    
    \draw[dashed, <->, thick] (A) -- (C);
    \draw[dashed, <->, thick] (B) -- (D);
    
    \node at (-0.5, 0) {};
    \node at (7.5, 0) {};
\end{tikzpicture}
\end{center}

Here we introduce generalized consistency training (GCT), which extends consistency training to any conditional ODE or forward diffusion process (through the corresponding conditional ODE). Intuitively, generalized consistency training modified consistency distillation (Eq. \ref{eq:L_CM} and Eq. \ref{eq:CD_sample}) in the same manner as how conditional flow matching modified the flow matching objective:
instead of requiring the unconditional ODE vector field $\mathbf{u}_t(\mathbf{x})$ which is not available without a pretrained diffusion or conditional flow matching model, GCT only requires the user-specified conditional ODE vector field $\mathbf{u}_t(\mathbf{x}\mid \mathbf{x}_0)$.
\vskip -0.3in
\begin{equation}\label{eq:gct}
    \begin{aligned}
    &\mathcal{L}_{\mathrm{GCT}}(\theta)\\
    &\quad=\mathbb{E}\left\|\left( \mathbf{c}_{\theta}(\mathbf{x}_{t_{i+1}},t_{i+1})-\text{stopgrad}\left(\mathbf{c}_{\theta} (\mathbf{x}_{t_{i}},t_{i}) \right)\right)\right\|, \\ &\qquad\qquad\qquad\mathbf{x}_{t_{i+1}}-\mathbf{x}_{t_{i}} = \mathbf{u}_{t_{i+1}}(\mathbf{x}_{t_{i+1}}\mid \mathbf{x}_0)(t_{i+1}-t_i)
        \end{aligned}
\end{equation}
Where the expectation is taken over $i$, $p(\mathbf{x}_0)$, and $p(\mathbf{x}_{t_{i+1}}|\mathbf{x}_0)$. An alternative formulation where the conditional flow is defined by $\mathbf{u}_{t_{i+1}}(\mathbf{x}\mid \mathbf{x}_0, \mathbf{x}_1)$ is detailed in Appendix \ref{appendix:alternative}.

 We proved that the generalized consistency training (GCT) objective is equivalent to the consistency distillation (CD) objective (Eq. \ref{eq:L_CM}, Eq. \ref{eq:CD_sample}). The proof is provided in Appendix \ref{appendix:gct}.

\begin{theorem}\label{theorem:GCT}
   Assuming the consistency function $\mathbf{c}_{\theta}(\mathbf{x},t)$ is twice differentiable with bounded second derivatives, and $\mathbb{E}_{p(\mathbf{x}_{0},\mathbf{x}_{1}\mid \mathbf{x})}\left[\left\|\mathbf{u}_{t}(\mathbf{x}\mid \mathbf{x}_{0},\mathbf{x}_{1})\right\|\right]<\infty$, up to a constant independent of $\theta$, $\mathcal{L}_{\mathrm{GCT}}$ and $ \mathcal{L}_{\mathrm{CD}}$ are equal.
\end{theorem}

\subsubsection{Inverse Consistency Models}

With generalized consistency training, we can now provide the inverse consistency model (ICM) (Figure \ref{fig:schema}, Algorithm \ref{alg:ICM}):
\begin{equation}\label{eq:L_ICM_v1}
    \begin{aligned}
    &\mathcal{L}_{\mathrm{ICM}}(\theta)\\
    &\quad=\mathbb{E}\left\|\left( \mathbf{c}_{\theta}(\mathbf{x}_{t_{i+1}},t_{i+1})-\text{stopgrad}\left(\mathbf{c}_{\theta} (\mathbf{x}_{t_{i}},t_{i}) \right) \right) \right\|, \\
    &\qquad\qquad\qquad\mathbf{x}_{t_{i+1}}-\mathbf{x}_{t_{i}} = \mathbf{u}_{t_{i+1}}(\mathbf{x}_{t_{i+1}}\mid \mathbf{x}_0)(t_{i+1}-t_i)
        \end{aligned}
\end{equation}
which is the consistency model counterpart of IFM (Eq. \ref{eq:L_IFM_v1}). The expectation is taken over $i$, $p(\mathbf{x}_1)$, $p\left(\mathbf{x}_{t_{i+1}} \mid \mathbf{x}_0=\mathbf{c}_{\theta} (\mathbf{x}_1,1)\right)$. Similar to IFM, a convenient alternative form is provided in Appendix \ref{appendix:alternative}. 

Since learning a consistency model is equivalent to learning a conditional flow matching model, ICM is equivalent to IFM following directly from our Theorem \ref{theorem:GCT} and Theorem 1 from \cite{songConsistencyModels2023}. 

\begin{lemma}\label{lemma:ICM}
   Assuming the consistency function $\mathbf{c}_{\theta}(\mathbf{x},t)$ is twice differentiable and $\partial\mathbf{c}_{\theta}(\mathbf{x},t)/\partial \mathbf{x}$ is almost everywhere nonzero\footnote{$\partial\mathbf{c}_{\theta}(\mathbf{x},t)/\partial \mathbf{x}\neq0$ is required to ensure the existence of corresponding ODE,  and it excludes trivial solution such as $\mathbf{c}_{\theta}(\mathbf{x},t)\equiv\mathrm{constant}$.  With identity initialization of $\mathbf{c}_{\theta}(\mathbf{x},t)$, we do not find it to be necessary for enforcing this condition in practice.}, when the inverse consistency loss $\mathcal{L}_{\mathrm{ICM}}=0$, there exists a corresponding ODE vector field $\mathbf{v}_{t}^{\theta}(x)$ that minimized the inverse flow matching loss $\mathcal{L}_{\mathrm{IFM}}$ to $0$.
\end{lemma}

The proof is provided in Appendix \ref{appendix:lemma2}. As in IFM, when the loss converges, the data distribution  $q\left(\mathbf{x}_0\right)$ recovered by ICM matches the ground truth distribution  $p(\mathbf{x}_0)$, but ICM is much more computationally efficient as it is a simulation-free objective.

\section{Experiments}\label{section:exp}
We first demonstrated the performance and properties of IFM and ICM on synthetic inverse generation datasets, which include a deterministic problem of inverting Naiver-Stokes simulation and a stochastic problem of denoising a synthetic noise dataset 8-gaussians. Next, we demonstrated that our method outperforms prior methods \citep{makinenCollaborativeFilteringCorrelated2020,krull_noise2void_2019,batsonNoise2SelfBlindDenoising2019} with the same neural network architecture on a semi-synthetic dataset of natural images with three synthetic noise types, and a real-world dataset of fluorescence microscopy images. Finally, we demonstrated that our method can be applied to denoise single-cell genomics data.

\begin{figure}[ht]
    \centering
    \includegraphics[width=0.45\textwidth]{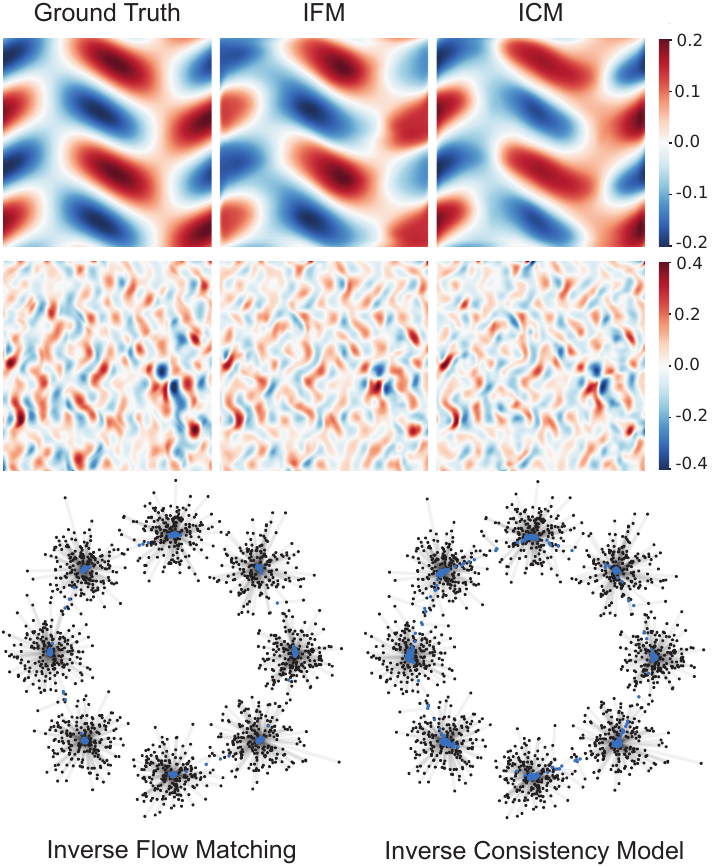}
      \vskip -0.1in
      \caption{Demonstration of inverse flow algorithms on synthetic datasets.
      Top panel shows an application to inverting Navier-Stokes fluid dynamics simulation color indicating the difference between the input state and the initial state. Bottom panel shows a denoising application on 8-gaussians dataset with input (black) and denoised data (blue) connected with lines.}
    \label{fig:toy}
      \vskip -0.2in
\end{figure}

\subsection{Synthetic datasets}\label{section:syn_datasets}

To test the capability of inverse flow in inverting complex transformations, we first attempted the deterministic inverse generation problem of inverting the transformation by Navier-Stokes fluid dynamics simulation\footnote{Inverse flow algorithms can be applied to deterministic transformations from $\mathbf{x}_0$ to $\mathbf{x}_1$ by using a matching conditional ODE, even though the general forms consider stochastic transforms described by $p(\mathbf{x}_1 \mid \mathbf{x}_0)$.}. We aim to recover the earlier state of the system without providing them for training (Figure \ref{fig:toy}). Navier-Stokes equations describe the motion of fluids by modeling the relationship between fluid velocity, pressure, viscosity, and external forces. These equations are fundamental in fluid dynamics and remain mathematically challenging, particularly in understanding turbulent flows.   The details of the simulation are described in Appendix \ref{appendix:syn_datasets}.

To test inverse flow algorithms on a denoising inverse generation problem, we generated a synthetic 8-gaussians dataset (Appendix \ref{appendix:syn_datasets} for details). Both IFM and ICM are capable of noise removal (Figure \ref{fig:toy}). ICM achieved a similar denoising performance as IFM, even though it is much more computationally efficient due to the iterative evaluation of ODE (NFE=10) by IFM.

\subsection{Semi-synthetic datasets}

We evaluated the proposed method on images in the benchmark dataset BSDS500 \citep{arbelaezContourDetectionHierarchical2011}, Kodak, and Set12 \citep{zhangGaussianDenoiserResidual2017}. To test the model's capability to deal with various types of conditional noise distribution, we generated synthetic noisy images for three different types of noise, including correlated noise and adding noise through a diffusion process without a closed-form transition density function (Appendix \ref{appendix:realdata} for details). All models were trained using the BSDS500 training set and evaluated on the BSDS500 test set, Kodak, and Set12. We show additional qualitative results in Appendix \ref{appendix:qualitative}.

\begin{table*}[]\label{table:qualitative}
\centering
\caption{Quantitative benchmark of denoising performances in multiple datasets for various noise distributions measured by Peak signal-to-noise ratio (PSNR) in dB.}
\tabcolsep=0.09cm
\begin{tabular}{cccccccc}
\hline
\multicolumn{2}{c}{Noise type}                  & Input & Supervised     & BM3D  & Noise2Void & Noise2Self & \textbf{Ours (ICM)} \\ \hline
\multirow{3}{*}{Gaussian}             & BSDS500 & 20.17 & 28.00          & 27.49 & 26.54      & 27.79      & \textbf{28.16}      \\
                                      & Kodak   & 20.18 & 28.91          & 28.54 & 27.55      & 28.72      & \textbf{29.08}      \\
                                      & Set12   & 20.16 & 28.99          & 28.95 & 27.79      & 28.78      & \textbf{29.19}      \\ \hline
\multirow{3}{*}{Correlated}           & BSDS500 & 20.17 & 27.10          & 24.48 & 26.32      & 21.03      & \textbf{27.64}      \\
                                      & Kodak   & 20.17 & 27.97          & 25.03 & 27.39      & 21.56      & \textbf{28.53}      \\
                                      & Set12   & 20.18 & 27.88          & 25.21 & 27.43      & 21.58      & \textbf{28.46}      \\ \hline
\multirow{3}{*}{SDE (Jacobi process)} & BSDS500 & 14.90 & \textbf{24.34} & 20.32 & 23.56      & 22.60      & 24.28               \\
                                      & Kodak   & 14.76 & \textbf{25.34} & 20.42 & 23.99      & 23.70      & 25.07               \\
                                      & Set12   & 14.80 & \textbf{25.01} & 20.51 & 24.43      & 23.26      & 24.74               \\ \hline
\end{tabular}
\end{table*}

\begin{enumerate}
\item{Gaussian noise:}
we added independent Gaussian noise with fixed variance.

\item{Correlated noise:}
we employed convolution kernels to generate correlated Gaussian noise following the method in \cite{makinenCollaborativeFilteringCorrelated2020}
\begin{equation}
    \begin{aligned}
        \eta=\nu \circledast g
    \end{aligned}
\end{equation}
where $\nu\sim \mathcal{N}(0, \sigma^{2}I)$ and $g$ is a convolution kernel.

\item{Jacobi process:}
we transformed the data with Jacobi process (Wright-Fisher diffusion), as an example of SDE-based transform without closed-form conditional distribution
\begin{equation}\label{eq:jacobi_diffusion}
\mathrm{d}\mathbf{x} = \frac{s}{2}[a(1-\mathbf{x})-b\mathbf{x}]\mathrm{d}t + \sqrt{s\mathbf{x}(1-\mathbf{x})} \mathrm{d}\mathbf{w}.
\end{equation}
 We generated corresponding noise data by simulating the Jacobi process with $s=1$ and $a=b=1$. Notably, the conditional noise distribution generated by the Jacobi process does not generally has an expectation that equals the ground truth (i.e. non-centered noise), which violates the assumptions of Noise2X methods.
\end{enumerate}

\begin{figure*}[h]
\vskip -0.1in
    \centering
    \includegraphics[width=0.65\textwidth]{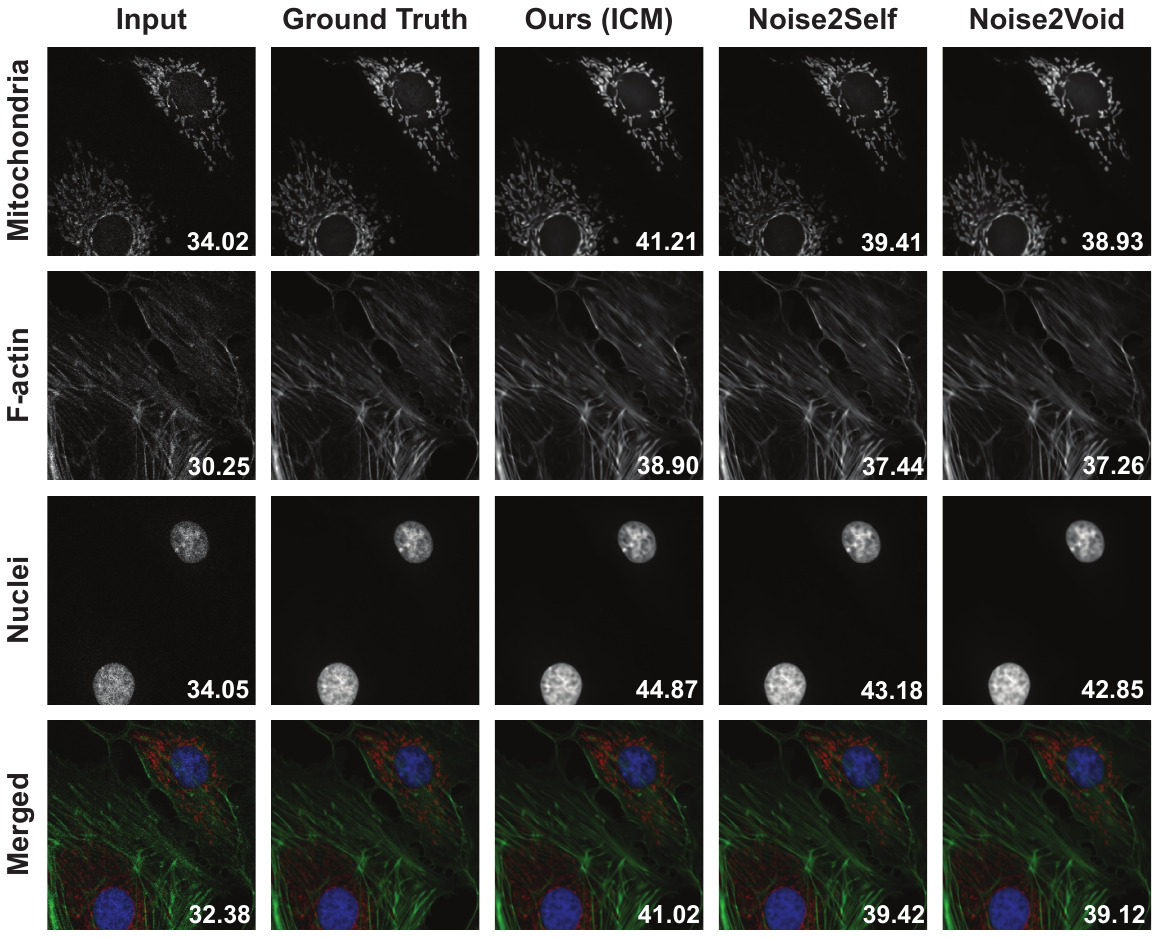}
        \vskip -0.1in
    \caption{Denoising results for fluorescence microscopy images with PSNR labelled.}
    \label{fig:qualitative}
      \vskip -0.2in

\end{figure*}

Our approach outperformed alternative unsupervised methods in all three noise types, especially in correlated noise and Jacobi process (Appendix \ref{appendix:qualitative}, Table \ref{table:qualitative}). This can be attributed to the fact that both Noise2X methods assumes independence of noise across different feature dimensions as well as centered-noise which were violated in corrleated noise and Jacobi process respectively.

Moreover, Our approach outperformed the supervised method on both Gaussian noise and correlated noise. Further analysis revealed that the supervised method encountered overfitting during the training process, which led to suboptimal performance. In contrast, our method did not exhibit such issues, highlighting the superiority of our approach.

In addition, in Appendix \ref{appendix:addition}, we conducted a series of experiments that demonstrate the reliability of our method under different intensities and types of noise. Furthermore, our method yielded satisfactory results even when there is a bias in the estimation of noise intensity. It also achieved excellent performance on RGB images and small sample-size datasets.

\subsection{Real-world datasets}

\subsubsection{Fluorescence Microscopy data (FMD)}

Fluorescence microscopy is an important scientific application of denoising without ground truth. Experimental constraints such as phototoxicity and frame rates often limit the capability to obtain clean data. We denoised confocal microscopy images from Fluorescence Microscopy Denoising (FMD) dataset \citep{zhang_poisson-gaussian_2019}. We first fitted a signal-dependent Poisson-Gaussian noise model adopted from \cite{liu_estimation_2013} for separate channels of each noisy microscopic images (Appendix \ref{appendix:micro} for details). Then denoising flow models were trained with the conditional ODE specified to be consistent with fitted noise model. Our method outperforms Noise2Self and Noise2Void, achieving superior denoising performance for mitochondria, F-actin, and nuclei in the microscopic images of BPAE cells (Figure \ref{fig:qualitative}).

\begin{figure*}[ht]
    \centering
          \vskip -0.1in
    \includegraphics[width=0.75\textwidth]{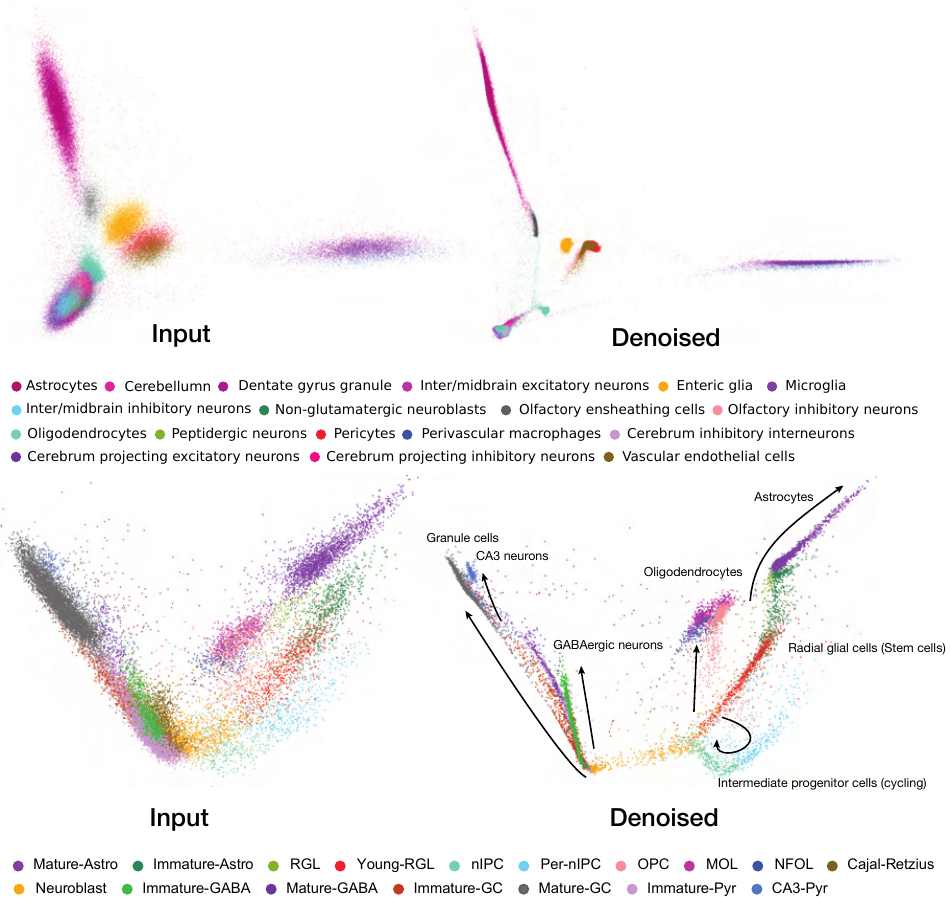}
    \vskip -0.05in
    \caption{Denoising single-cell RNA-seq data with ICM improves resolution for cell types and developmental trajectories. The top two principal components are visualized. Top panel: results for \cite{zeisel_molecular_2018}. Bottom panel: results for \cite{hochgerner_conserved_2018}, Astro: astrocytes, RGL: radial glial cells, IPC: intermediate progenitor cells, OPC: oligodendrocyte precursor cells, MOL: mature oligodendrocytes; NFOL: newly formed oligodendrocytes, GABA: GABAergic neurons, GC: granule cells, Pyr: pyramidal neurons.}
    \label{fig:scRNA}
    \vskip -0.18in
\end{figure*}

\subsubsection{Application to denoise single-cell genomics data}\label{section:scRNA}

In recent years, the development of single-cell sequencing technologies has enabled researchers to obtain more fine-grained information on tissues and organs at the resolution of single cells. However,  the low amount of sample materials per-cell introduces considerable noise in single-cell genomics data. These noises may obscure real biological signals, thereby affecting subsequent analyses.

Applying ICM to an adult mouse brain single-cell RNA-seq dataset \citep{zeisel_molecular_2018} and a mouse brain development single-cell RNA-seq dataset \citep{hochgerner_conserved_2018} (Figure \ref{fig:scRNA}, Appendix \ref{appendix:scRNA} for details), we observed that the denoised data better reflects the cell types and developmental trajectories. We compared the original and denoised data by the accuracy of predicting the cell type identity of each cell based on its nearest neighbor in the top two principal components. Our methods improved the accuracy of the adult mouse brain dataset from $0.513\pm0.003$ to $0.571\pm0.003$, and the mouse brain development dataset from 
$0.647\pm0.006$ to $0.736\pm0.006$.

\section{Limitation and Conclusion}\label{section:conclusion}
We introduce Inverse Flow (IF), a generative modeling framework for inverse generation problems such as denoising without ground truth, and two methods Inverse Flow Match (IFM) and Inverse Consistency Model (ICM) to solve the inverse flow problem. Our framework connects the family of continuous-time generative models to inverse generation problems. Practically, we extended the applicability of denoising without ground truth to almost any continuous noise distributions.  We demonstrated strong empirical results applying inverse flow. A limitation of inverse flow is assuming prior knowledge of the noise distribution, and future work is needed to relax this assumption. We expect inverse flow to open up possibilities to explore additional connections to the expanding family of continuous-time generative model methods, and the generalized consistency training objective will expand the application of consistency models.

\bibliography{InverseFlow}
\bibliographystyle{iclr2025_conference}

\newpage
\onecolumn
\appendix

\section{Appendix}

\subsection{Alternative forms of IFM and ICM}\label{appendix:alternative}
Here we provide the details of alternative objectives and corresponding algorithms of IFM and ICM which are easier and flexible to use.

\subsubsection{Alternative objectives of IFM and ICM}

We define the alternative objective of IFM similar to conditional flow matching (Eq. \ref{eq:L_CFM_v2}):
\begin{equation}\label{eq:L_IFM_v2}
    \begin{aligned}
    \mathcal{L}_{\mathrm{IFM}}(\theta)=\mathbb{E}_{t, p(\mathbf{x}_1), p\left(\mathbf{x}_{1}'\mid\mathbf{x}_0 = \textrm{ODE}^{\mathbf{v}^{\theta}}_{1\rightarrow 0}(\mathbf{x}_1)\right),p(\mathbf{x}_{t}\mid\mathbf{x}_{0},\mathbf{x}_{1}')}\left[\left\|\mathbf{v}^{\theta}_t(\mathbf{x}_{t})-\mathbf{u}_t\left(\mathbf{x}_{t} \mid \textrm{ODE}^{\mathbf{v}^{\theta}}_{1\rightarrow 0}(\mathbf{x}_1), \mathbf{x}'_1\right)\right\|\right]
        \end{aligned}
\end{equation}
where $\mathbf{x}'_1$ is sampled from the conditional noise distribution. As described in Section \ref{section:cfm} $\mathbf{u}_t\left(\mathbf{x} \mid \mathbf{x}_0, \mathbf{x}_1'\right)$ can be easily chosen as any smooth interpolation between $\mathbf{x}_0$ and $\mathbf{x}_1'$, such as  $\mathbf{u}_t\left(\mathbf{x} \mid \mathbf{x}_0, \mathbf{x}_1'\right) = \mathbf{x}_1' - \mathbf{x}_0$.

Since ICM is based on generalized consistency training, we first provide the alternative objective of generalized consistency training
\begin{equation}\label{eq:gct_2}
    \begin{aligned}
    \mathcal{L}_{\mathrm{GCT}}(\theta)=\mathbb{E}_{i, p(\mathbf{x}_0,\mathbf{x}_1),p(\mathbf{x}_{t_{i+1}}|\mathbf{x}_0, \mathbf{x}_1)}\left[ \left\|\mathbf{c}_{\theta}(\mathbf{x}_{t_{i+1}},t_{i+1})-\text{stopgrad}\left(\mathbf{c}_{\theta} (\mathbf{x}_{t_{i}},t_{i}) \right)\right\| \right], \\ \mathbf{x}_{t_{i+1}}-\mathbf{x}_{t_{i}} = \mathbf{u}_{t_{i+1}}(\mathbf{x}_{t_{i+1}}\mid \mathbf{x}_0, \mathbf{x}_1)(t_{i+1}-t_i)
        \end{aligned}
\end{equation}
where the conditional flow is defined jointly by $p(\mathbf{x}_1\mid\mathbf{x}_0)$ and $\mathbf{u}_{t_{i+1}}(\mathbf{x}\mid \mathbf{x}_0, \mathbf{x}_1)$.

Then the alterntive form of ICM can be defined as 
\begin{equation}\label{eq:L_ICM_v2}
    \begin{aligned}
    &\mathcal{L}_{\mathrm{ICM}}(\theta) = \\
    &\quad \mathbb{E}_{i, p(\mathbf{x}_1), p\left(\mathbf{x}'_1 \mid \mathbf{x}_0=\mathbf{c}_{\theta} (\mathbf{x}_1,1)\right), p\left(\mathbf{x}_{t_{i+1}} \mid \mathbf{x}_0=\mathbf{c}_{\theta} (\mathbf{x}_1,1), \mathbf{x}'_1\right)}\left[ \left\| \mathbf{c}_{\theta}(\mathbf{x}_{t_{i+1}},t_{i+1})-\text{stopgrad}\left(\mathbf{c}_{\theta} (\mathbf{x}_{t_{i}},t_{i}) \right) \right\| \right], \\
    &\qquad\qquad\qquad\qquad\qquad\qquad\qquad\qquad\qquad\qquad\mathbf{x}_{t_{i+1}}-\mathbf{x}_{t_{i}} = \mathbf{u}_{t_{i+1}}(\mathbf{x}_{t_{i+1}}\mid \mathbf{x}_0,\mathbf{x}'_1)(t_{i+1}-t_i)
        \end{aligned}
\end{equation}

where $\mathbf{u}_{t}(\mathbf{x}\mid \mathbf{x}_0, \mathbf{x}_1')$ can be freely defined based on any interpolation between $\mathbf{x}_0$ and $\mathbf{x}_1'$, which is more easily applicable to any conditional noise distribution:.

\subsubsection{Alternative algorithms of IFM and ICM}\label{appendix:alg}
Here we show the algorithms of alternative objectives of IFM (Eq. \ref{eq:L_IFM_v2}) and ICM (Eq. \ref{eq:L_ICM_v2}).
\begin{figure}[H]
\begin{minipage}[H]{0.45\textwidth}
\begin{algorithm}[H]
    \centering
    \caption{IFM Training v2.}\label{alg:IFM_v2}
    \begin{algorithmic}[1]
        \State \textbf{Input:} dataset $\mathcal{D}$, initial model parameter $\theta$, and learning rate $\eta$
        \Repeat
        \State Sample $\mathbf{x}_{1}\sim \mathcal{D}$ and  $t\sim \mathcal{U}[0,1]$
        \State $\mathbf{x}_0 \leftarrow \textrm{stopgrad}\left(\textrm{ODE}^{\mathbf{v}^{\theta}}_{1\rightarrow 0}(\mathbf{x}_1)\right)$
        \State Sample $\mathbf{x}_{1}'\sim p(\mathbf{x}_{1}'\mid \mathbf{x}_{0})$
        \State Sample $\mathbf{x}_{t}\sim p(\mathbf{x}_{t}\mid \mathbf{x}_{0},\mathbf{x}_{1}')$
        \State $\begin{multlined}
            \mathcal{L}(\theta)\leftarrow \\
            \left\|\mathbf{v}^{\theta}_t(\mathbf{x}_{t})-\mathbf{u}_t\left(\mathbf{x}_{t} \mid \mathbf{x}_{0},\mathbf{x}_{1}'\right)\right\|^2
        \end{multlined}$
        \State  $\theta\leftarrow\theta-\eta\nabla_{\theta}\mathcal{L}(\theta)$
        \Until {convergence}
    \end{algorithmic}
\end{algorithm}
\end{minipage}
\begin{minipage}[H]{0.53\textwidth}
\begin{algorithm}[H]
    \centering
    \caption{ICM Training v2.}\label{alg:ICM_v2}
    \begin{algorithmic}[1]
        \State \textbf{Input:} dataset $\mathcal{D}$, initial model parameter $\theta$, learning rate $\eta$, and sequence of time points $0=t_{1}<t_{2}<\dots<t_{N}=1$
        \Repeat
        \State Sample $\mathbf{x}_{1}\sim \mathcal{D}$ and  $i\sim \mathcal{U}[1,N-1]$
        \State $\mathbf{x}_0 \leftarrow \textrm{stopgrad}\left(\mathbf{c}_{\theta}(\mathbf{x}_{1},1)\right)$
        \State Sample $\mathbf{x}_{1}'\sim p(\mathbf{x}_{1}'\mid \mathbf{x}_{0})$
        \State Sample $\mathbf{x}_{t_{i+1}}\sim p(\mathbf{x}_{t_{i+1}}\mid \mathbf{x}_{0},\mathbf{x}_{1}')$
        \State $\begin{multlined}
            \mathbf{x}_{t_{i}}\leftarrow \\
            \mathbf{x}_{t_{i+1}}-\mathbf{u}_{t_{i+1}}(\mathbf{x}_{t_{i+1}}\mid \mathbf{x}_0,\mathbf{x}_{1}')(t_{i+1}-t_i)
        \end{multlined}$
        \State $\begin{multlined}
            \mathcal{L}(\theta)\leftarrow \\
       d\left[ \mathbf{c}_{\theta}(\mathbf{x}_{t_{i+1}},t_{i+1}),\textrm{stopgrad}\left(\mathbf{c}_{\theta}(\mathbf{x}_{t_{i}},t_{i}) \right)\right]
        \end{multlined}$
        \State  $\theta\leftarrow\theta-\eta\nabla_{\theta}\mathcal{L}(\theta)$
        \Until {convergence}
    \end{algorithmic}
\end{algorithm}
\end{minipage}
\end{figure}

\subsection{Proofs}\label{appendix:proof}

\subsubsection{inverse flow Matching}\label{appendix:IFM}

\textbf{Theorem 1:} Assume that the conditional noise distribution $p(\mathbf{x}_{1}\mid \mathbf{x}_{0})$ satisfies the condition that,  for any noisy data distribution $p(\mathbf{x}_{1})$ there exists only one probability distribution $p(\mathbf{x}_{0})$ that satisfies $p(\mathbf{x}_{1})=\int p(\mathbf{x}_{1}\mid \mathbf{x}_{0})p(\mathbf{x}_{0})\mathrm{d}\mathbf{x}_0$ , then under the condition that $\mathcal{L}_{\mathrm{IFM}}=0$,  we have $q(\mathbf{x}_{0})=p(\mathbf{x}_{0})$.

\textit{Proof:}

The inferred data distribution is given by the push-forward operator \citep{lipmanFlowMatchingGenerative2023}:
\begin{equation}
    \begin{aligned}
q(\mathbf{x}_0)=\left[\textrm{ODE}^{\mathbf{v}^{\theta}}_{1\rightarrow 0}\right]*p(\mathbf{x}_{1})
    \end{aligned}
\end{equation}
which is defined for any continuous normalizing flow $\phi$ from $\mathbf{x}_{1}$ to $\mathbf{x}_{0}$ in the form of
\begin{equation}
    \begin{aligned}
        [\phi]*p(\mathbf{x}_{1})=p\left(\phi^{-1}(\mathbf{x}_{0})\right)\mathrm{det}\left[\frac{\partial\phi^{-1}}{\partial \mathbf{x}}(\mathbf{x}_{0})\right]
    \end{aligned}
\end{equation}
where $\mathbf{x}_{1}=\phi^{-1}(\mathbf{x}_{0})$.
The inferred noisy data distribution $q(\mathbf{x}_{1})$ is given by
\begin{equation}\label{eq:23}
    \begin{aligned}
        q(\mathbf{x}_{1})=\int p(\mathbf{x}_{1}\mid \mathbf{x}_{0})q(\mathbf{x}_{0})\mathrm{d}\mathbf{x}_{0}
    \end{aligned}
\end{equation}
Under the condition $\mathcal{L}_{\mathrm{IFM}}=0$, we have
\begin{equation}
    \begin{aligned}
        q(\mathbf{x}_0)=\left[\textrm{ODE}^{\mathbf{v}^{\theta}}_{1\rightarrow 0}\right]*q(\mathbf{x}_{1})
    \end{aligned}
\end{equation}
Then we find that
\begin{equation}
    \begin{aligned}
        \left[\textrm{ODE}^{\mathbf{v}^{\theta}}_{1\rightarrow 0}\right]*p(\mathbf{x}_{1})=\left[\textrm{ODE}^{\mathbf{v}^{\theta}}_{1\rightarrow 0}\right]*q(\mathbf{x}_{1})
    \end{aligned}
\end{equation}
By the definition of the push-forward operator, we have
\begin{equation}
    \begin{aligned}
        p\left(\left(\textrm{ODE}^{\mathbf{v}^{\theta}}_{1\rightarrow 0}\right)^{-1}(\mathbf{x}_{0})\right)\mathrm{det}\left[\frac{\partial \left(\textrm{ODE}^{\mathbf{v}^{\theta}}_{1\rightarrow 0}\right)^{-1}}{\partial \mathbf{x}}(\mathbf{x}_{0})\right] \\
        = q\left(\left(\textrm{ODE}^{\mathbf{v}^{\theta}}_{1\rightarrow 0}\right)^{-1}(\mathbf{x}_{0})\right)\mathrm{det}\left[\frac{\partial \left(\textrm{ODE}^{\mathbf{v}^{\theta}}_{1\rightarrow 0}\right)^{-1}}{\partial \mathbf{x}}(\mathbf{x}_{0})\right]
    \end{aligned}
\end{equation}
Since the solution of  ODE is unique, $\textrm{ODE}^{\mathbf{v}^{\theta}}_{1\rightarrow 0}$ is a bijective function with
$$\left(\textrm{ODE}^{\mathbf{v}^{\theta}}_{1\rightarrow 0}\right)^{-1}=\textrm{ODE}^{\mathbf{v}^{\theta}}_{0\rightarrow 1}$$
and
$$\mathbf{x}_{1}=\textrm{ODE}^{\mathbf{v}^{\theta}}_{0\rightarrow 1}(\mathbf{x}_{0})=\left(\textrm{ODE}^{\mathbf{v}^{\theta}}_{1\rightarrow 0}\right)^{-1}(\mathbf{x}_{0})$$
Also, the nontrivial solution ensures that the determinant is non-zero. By substitution, we get
\begin{equation}
    \begin{aligned}
        p(\mathbf{x}_{1})=q(\mathbf{x}_{1})
    \end{aligned}
\end{equation}
and combine with Eq. \ref{eq:23}, we find that
\begin{equation}
    \begin{aligned}
        p(\mathbf{x}_{1})=\int p(\mathbf{x}_{1}\mid \mathbf{x}_{0})q(\mathbf{x}_{0})\mathrm{d}\mathbf{x}_{0}
    \end{aligned}
\end{equation}
We close the proof by directly applying the uniqueness of $p(\mathbf{x}_{0})$ and find that
\begin{equation}
    \begin{aligned}
        q(\mathbf{x}_{0})=p(\mathbf{x}_{0})
    \end{aligned}
\end{equation}

\textbf{Remark 1:} Our proof utilizes the one-to-one mapping property of neural ODEs \citep{kidger_neural_2022}, which guarantees a diffeomorphic map between the noisy data distribution and the inferred data distribution.

\textbf{Remark 2:} Many inverse problems are ill-posed and impossible to address perfectly. The assumptions in our Theorem 1 provides guidance on the conditions under which IFM can recover the ground truth distribution. These conditions are:
\begin{itemize}
    \item Knowledge of the noisy data distribution: Either directly or through access to sufficient noisy data.
    \item Transformability of $p(\mathbf{x}_1)$: The noisy data distribution $p(\mathbf{x}_1)$ must be transformable from $p(\mathbf{x}_0)$ via an ODE. This condition accommodates nearly any continuous noise distribution but excludes ill-posed transformations that lack a one-on-one mapping between $p(\mathbf{x}_0)$ and $p(\mathbf{x}_1)$ (e.g., transformations like converting color images to grayscale).
\end{itemize}

\textbf{Lemma 1:}
    Given a conditional ODE vector field $\mathbf{u}_{t}(\mathbf{x}\mid \mathbf{x}_{0}, \mathbf{x}_{1})$ that generates a conditional probability path $p(\mathbf{x}_{t}\mid \mathbf{x}_{0}, \mathbf{x}_{1})$, the unconditional probability path $p(\mathbf{x}_{t})$ can be generated by the unconditional ODE vector field $\mathbf{u}_{t}(\mathbf{x})$, which is defined as
\begin{equation}
    \begin{aligned}
        \mathbf{u}_{t}(\mathbf{x})=\mathbb{E}_{p(\mathbf{x}_{0},\mathbf{x}_{1}\mid \mathbf{x})}\left[\mathbf{u}_{t}(\mathbf{x}\mid \mathbf{x}_{0},\mathbf{x}_{1})\right]
    \end{aligned}
\end{equation}

\textit{Proof:}

To verify this, we check that $p(\mathbf{x}_{t})$ and $\mathbf{u}_{t}(\mathbf{x})$ satisfy the continuity equation:
\begin{equation}
    \begin{aligned}
        \frac{d}{dt}p(\mathbf{x}_{t})+\mathrm{div}\left(\mathbf{u}_{t}(\mathbf{x})p(\mathbf{x}_{t})\right)=0.
    \end{aligned}
\end{equation}

By definition,
\begin{equation}
    \begin{aligned}
        \frac{d}{dt}p(\mathbf{x}_{t})=\frac{d}{dt}\int p(\mathbf{x}_{t}|\mathbf{x}_{0},\mathbf{x}_{1})p(\mathbf{x}_{0}, \mathbf{x}_{1})\mathrm{d}\mathbf{x}_{0}\mathrm{d}\mathbf{x}_{1}.
    \end{aligned}
\end{equation}

With Leibniz Rule we have
\begin{equation}
    \begin{aligned}\label{eq:Lemma}
        \frac{d}{dt}p(\mathbf{x}_{t})=\int \frac{d}{dt}p(\mathbf{x}_{t}|\mathbf{x}_{0},\mathbf{x}_{1})p(\mathbf{x}_{0}, \mathbf{x}_{1})\mathrm{d}\mathbf{x}_{0}\mathrm{d}\mathbf{x}_{1}.
    \end{aligned}
\end{equation}
Since $\mathbf{u}_{t}(\mathbf{x}|\mathbf{x}_{0}, \mathbf{x}_{1})$ generates $p(\mathbf{x}_{t}|\mathbf{x}_{0},\mathbf{x}_{1})$, by the continuity equation we have
\begin{equation}
    \begin{aligned}
        \frac{d}{dt}p(\mathbf{x}_{t}|\mathbf{x}_{0},\mathbf{x}_{1})+\mathrm{div}\left(\mathbf{u}_{t}(\mathbf{x}|\mathbf{x}_{0}, \mathbf{x}_{1})p(\mathbf{x}_{t}|\mathbf{x}_{0},\mathbf{x}_{1})\right)=0.
    \end{aligned}
\end{equation}

Substitution in Eq. \ref{eq:Lemma} gives
\begin{equation}
    \begin{aligned}
        \frac{d}{dt}p(\mathbf{x}_{t})=-\int \mathrm{div}\left(\mathbf{u}_{t}(\mathbf{x}|\mathbf{x}_{0}, \mathbf{x}_{1})p(\mathbf{x}_{t}|\mathbf{x}_{0},\mathbf{x}_{1})\right)p(\mathbf{x}_{0}, \mathbf{x}_{1})\mathrm{d}\mathbf{x}_{0}\mathrm{d}\mathbf{x}_{1}.
    \end{aligned}
\end{equation}

Exchanging the derivative and integral,
\begin{equation}
    \begin{aligned}\label{eq:deint}
        \frac{d}{dt}p(\mathbf{x}_{t})=-\mathrm{div}\int\mathbf{u}_{t}(\mathbf{x}|\mathbf{x}_{0}, \mathbf{x}_{1}) p(\mathbf{x}_{t}|\mathbf{x}_{0},\mathbf{x}_{1})p(\mathbf{x}_{0}, \mathbf{x}_{1})\mathrm{d}\mathbf{x}_{0}\mathrm{d}\mathbf{x}_{1}.
    \end{aligned}
\end{equation}

The definition of $\mathbf{u}_{t}(\mathbf{x})$ is
\begin{equation}
    \begin{aligned}\label{eq:def_u}
        \mathbf{u}_{t}(\mathbf{x})=\mathbb{E}_{p(\mathbf{x}_{0},\mathbf{x}_{1}\mid \mathbf{x})}\left[\mathbf{u}_{t}(\mathbf{x}\mid \mathbf{x}_{0},\mathbf{x}_{1})\right]=\int \mathbf{u}_{t}(\mathbf{x}\mid \mathbf{x}_{0},\mathbf{x}_{1})\frac{p(\mathbf{x}_{t}|\mathbf{x}_{0},\mathbf{x}_{1})p(\mathbf{x}_{0}, \mathbf{x}_{1})}{p(\mathbf{x}_{t})}\mathrm{d}\mathbf{x}_{0}\mathrm{d}\mathbf{x}_{1}.
    \end{aligned}
\end{equation}

Combining Eq. \ref{eq:deint} and Eq. \ref{eq:def_u} gives the continuity equation:
\begin{equation}
    \begin{aligned}
        \frac{d}{dt}p(\mathbf{x}_{t})+\mathrm{div}\left(\mathbf{u}_{t}(\mathbf{x})p(\mathbf{x}_{t})\right)=0.
    \end{aligned}
\end{equation}

\subsubsection{Generalized Consistency Training}\label{appendix:gct}

Without loss of generality, we provide the proof for the form of $\mathcal{L}_{\mathrm{GCT}}$ in Eq. \ref{eq:gct_2}, and the proof for the form Eq. \ref{eq:gct} follows by assuming that the forward conditional probability path is independent of $\mathbf{x}_1$.

\textbf{Theorem 2}: Assuming the consistency function $\mathbf{c}_{\theta}(\mathbf{x},t)$ is twice differentiable with bounded second derivatives, and $\mathbb{E}_{p(\mathbf{x}_{0},\mathbf{x}_{1}\mid \mathbf{x})}\left[\left\|\mathbf{u}_{t}(\mathbf{x}\mid \mathbf{x}_{0},\mathbf{x}_{1})\right\|\right]<\infty$, up to a constant independent of $\theta$, $\mathcal{L}_{\mathrm{GCT}}$ and $ \mathcal{L}_{\mathrm{CD}}$ are equal.

\textit{Proof:}

The proof is inspired by \cite{songConsistencyModels2023}. We use the shorthand $\mathbf{c}_{\theta^{-}}$ to denote the stopgrad version of the consistency function $\mathbf{c}$. Given a multi-variate function $\mathbf{h}(\mathbf{x},\mathbf{y})$, the operator $\partial_{1}\mathbf{h}(\mathbf{x},\mathbf{y})$ and $\partial_{2}\mathbf{h}(\mathbf{x},\mathbf{y})$ denote the partial derivative with respect to $\mathbf{x}$ and $\mathbf{y}$. Let $\Delta t:=\mathrm{max}_{i}\left\{\mid t_{i+1}-t_{i}\mid\right\}$ and we use $o(\Delta t)$ to denote infinitesimal with respect to $\Delta t$.

Based on Eq. \ref{eq:CD_sample} and Eq. \ref{eq:L_CM}, the consistency distillation objective is
\begin{equation}
    \begin{aligned}
        \mathcal{L}_{\mathrm{CD}}(\theta)=\mathbb{E}_{i, p(\mathbf{x}_{0},\mathbf{x}_{1}),p(\mathbf{x}_{t_{i+1}}\mid \mathbf{x}_{0},\mathbf{x}_{1})}\left\{ d\left[ \mathbf{c}_\theta(\mathbf{x}_{t_{i+1}},t_{i+1}), \mathbf{c}_{\theta^{-}} (\mathbf{x}_{t_{i}},t_{i}) \right]  \right\}
    \end{aligned}
\end{equation}
where $\mathbf{x}_{t_{i}}=\mathbf{x}_{t_{i+1}}-(t_{i+1}-t_{i})\mathbf{u}_{t_{i+1}}(\mathbf{x}_{t_{i+1}})$ and $d$ is a general distance function.

We assume $d$ and $\mathbf{c}_{\theta^{-}}$ are twice continuously differentiable with bounded derivatives. With Taylor expansion, we have
\begin{equation}
    \begin{aligned}
        \mathcal{L}_{\mathrm{CD}}(\theta)&=\mathbb{E}_{i, p(\mathbf{x}_{0},\mathbf{x}_{1}),p(\mathbf{x}_{t_{i+1}}\mid \mathbf{x}_{0},\mathbf{x}_{1})}\left\{ d\left[ \mathbf{c}_{\theta}(\mathbf{x}_{t_{i+1}},t_{i+1}), \mathbf{c}_{\theta^{-}} (\mathbf{x}_{t_{i}},t_{i}) \right]  \right\} \\
        &=\mathbb{E}_{i, p(\mathbf{x}_{0},\mathbf{x}_{1}),p(\mathbf{x}_{t_{i+1}}\mid \mathbf{x}_{0},\mathbf{x}_{1})}\left\{ d\left[ \mathbf{c}_{\theta}(\mathbf{x}_{t_{i+1}},t_{i+1}), \mathbf{c}_{\theta^{-}} (\mathbf{x}_{t_{i+1}}-(t_{i+1}-t_{i})\mathbf{u}_{t_{i+1}}(\mathbf{x}_{t_{i+1}}),t_{i}) \right]  \right\} \\
        &=\mathbb{E}_{i, p(\mathbf{x}_{0},\mathbf{x}_{1}),p(\mathbf{x}_{t_{i+1}}\mid \mathbf{x}_{0},\mathbf{x}_{1})}\left\{ d\left[ \mathbf{c}_{\theta}(\mathbf{x}_{t_{i+1}},t_{i+1}), \mathbf{c}_{\theta^{-}} (\mathbf{x}_{t_{i+1}}, t_{i+1})\right. \right.\\
        &\quad-\partial_{1}\mathbf{c}_{\theta^{-}} (\mathbf{x}_{t_{i+1}},t_{i+1})(t_{i+1}-t_{i})\mathbf{u}_{t_{i+1}}(\mathbf{x}_{t_{i+1}}) \\
        &\quad\quad\left.\left.-\partial_{2}\mathbf{c}_{\theta^{-}} (\mathbf{x}_{t_{i+1}},t_{i+1})(t_{i+1}-t_{i}) +o(\Delta t)\right]  \right\}\\
        &=\mathbb{E}_{i, p(\mathbf{x}_{0},\mathbf{x}_{1}),p(\mathbf{x}_{t_{i+1}}\mid \mathbf{x}_{0},\mathbf{x}_{1})}\left\{ d\left[ \mathbf{c}_{\theta}(\mathbf{x}_{t_{i+1}},t_{i+1}), \mathbf{c}_{\theta^{-}} (\mathbf{x}_{t_{i+1}}, t_{i+1})\right] \right\} \\
        &\quad-\mathbb{E}_{i, p(\mathbf{x}_{0},\mathbf{x}_{1}),p(\mathbf{x}_{t_{i+1}}\mid \mathbf{x}_{0},\mathbf{x}_{1})}\left\{\partial_{2} d\left[ \mathbf{c}_{\theta}(\mathbf{x}_{t_{i+1}},t_{i+1}), \mathbf{c}_{\theta^{-}} (\mathbf{x}_{t_{i+1}}, t_{i+1})\right] \right.\\
        &\qquad\left.\cdot\left[\partial_{1}\mathbf{c}_{\theta^{-}} (\mathbf{x}_{t_{i+1}},t_{i+1})(t_{i+1}-t_{i})\mathbf{u}_{t_{i+1}}(\mathbf{x}_{t_{i+1}})\right] \right\}\\
        &\qquad\quad-\mathbb{E}_{i, p(\mathbf{x}_{0},\mathbf{x}_{1}),p(\mathbf{x}_{t_{i+1}}\mid \mathbf{x}_{0},\mathbf{x}_{1})}\left\{\partial_{2} d\left[ \mathbf{c}_{\theta}(\mathbf{x}_{t_{i+1}},t_{i+1}), \mathbf{c}_{\theta^{-}} (\mathbf{x}_{t_{i+1}}, t_{i+1})\right] \right.\\
        &\qquad\qquad\left.\cdot\left[\partial_{2}\mathbf{c}_{\theta^{-}} (\mathbf{x}_{t_{i+1}},t_{i+1})(t_{i+1}-t_{i}) \right]  \right\}+\mathbb{E}\left[o(\Delta t)\right]
    \end{aligned}
\end{equation}
Then, we apply Lemma \ref{lemma:1} and use Taylor expansion in the reverse direction,

\begingroup
\allowdisplaybreaks 
    \begin{align}
        &\mathcal{L}_{\mathrm{CD}}(\theta) \notag\\ 
        &=\mathbb{E}_{i, p(\mathbf{x}_{0},\mathbf{x}_{1}),p(\mathbf{x}_{t_{i+1}}\mid \mathbf{x}_{0},\mathbf{x}_{1})}\left\{ d\left[ \mathbf{c}_{\theta}(\mathbf{x}_{t_{i+1}},t_{i+1}), \mathbf{c}_{\theta^{-}} (\mathbf{x}_{t_{i+1}}, t_{i+1})\right] \right\} \notag\\
        &\quad-\mathbb{E}_{i, p(\mathbf{x}_{0},\mathbf{x}_{1}),p(\mathbf{x}_{t_{i+1}}\mid \mathbf{x}_{0},\mathbf{x}_{1})}\left\{\partial_{2} d\left[ \mathbf{c}_{\theta}(\mathbf{x}_{t_{i+1}},t_{i+1}), \mathbf{c}_{\theta^{-}} (\mathbf{x}_{t_{i+1}}, t_{i+1})\right] \right.\notag\\
        &\qquad\left.\cdot\left[\partial_{1}\mathbf{c}_{\theta^{-}} (\mathbf{x}_{t_{i+1}},t_{i+1})(t_{i+1}-t_{i})\mathbb{E}_{p(\mathbf{x}_{0},\mathbf{x}_{1} \mid \mathbf{x}_{t_{i+1}})}\left[\mathbf{u}_{t_{i+1}}(\mathbf{x}_{t_{i+1}}\mid \mathbf{x}_{0},\mathbf{x}_{1})\right] \right]\right\}\notag\\
        &\qquad\quad-\mathbb{E}_{i, p(\mathbf{x}_{0},\mathbf{x}_{1}),p(\mathbf{x}_{t_{i+1}}\mid \mathbf{x}_{0},\mathbf{x}_{1})}\left\{\partial_{2} d\left[ \mathbf{c}_{\theta}(\mathbf{x}_{t_{i+1}},t_{i+1}), \mathbf{c}_{\theta^{-}} (\mathbf{x}_{t_{i+1}}, t_{i+1})\right] \right.\notag\\
        &\qquad\qquad\left.\cdot\left[\partial_{2}\mathbf{c}_{\theta^{-}} (\mathbf{x}_{t_{i+1}},t_{i+1})(t_{i+1}-t_{i}) \right]  \right\}+\mathbb{E}\left[o(\Delta t)\right]\notag\\
        &\overset{(i)}{=}\mathbb{E}_{i, p(\mathbf{x}_{0},\mathbf{x}_{1}),p(\mathbf{x}_{t_{i+1}}\mid \mathbf{x}_{0},\mathbf{x}_{1})}\left\{ d\left[ \mathbf{c}_{\theta}(\mathbf{x}_{t_{i+1}},t_{i+1}), \mathbf{c}_{\theta^{-}} (\mathbf{x}_{t_{i+1}}, t_{i+1})\right] \right\} \notag\\
        &\quad-\mathbb{E}_{i, p(\mathbf{x}_{0},\mathbf{x}_{1}),p(\mathbf{x}_{t_{i+1}}\mid \mathbf{x}_{0},\mathbf{x}_{1})}\left\{\partial_{2} d\left[ \mathbf{c}_{\theta}(\mathbf{x}_{t_{i+1}},t_{i+1}), \mathbf{c}_{\theta^{-}} (\mathbf{x}_{t_{i+1}}, t_{i+1})\right] \right.\notag\\
        &\qquad\left.\cdot\left[\partial_{1}\mathbf{c}_{\theta^{-}} (\mathbf{x}_{t_{i+1}},t_{i+1})(t_{i+1}-t_{i})\mathbf{u}_{t_{i+1}}(\mathbf{x}_{t_{i+1}}\mid \mathbf{x}_{0},\mathbf{x}_{1}) \right]\right\}\notag\\
        &\qquad\quad-\mathbb{E}_{i, p(\mathbf{x}_{0},\mathbf{x}_{1}),p(\mathbf{x}_{t_{i+1}}\mid \mathbf{x}_{0},\mathbf{x}_{1})}\left\{\partial_{2} d\left[ \mathbf{c}_{\theta}(\mathbf{x}_{t_{i+1}},t_{i+1}), \mathbf{c}_{\theta^{-}} (\mathbf{x}_{t_{i+1}}, t_{i+1})\right] \right.\notag\\
        &\qquad\qquad\left.\cdot\left[\partial_{2}\mathbf{c}_{\theta^{-}} (\mathbf{x}_{t_{i+1}},t_{i+1})(t_{i+1}-t_{i}) \right]  \right\}+\mathbb{E}\left[o(\Delta t)\right]\notag\\
        &=\mathbb{E}_{i, p(\mathbf{x}_{0},\mathbf{x}_{1}),p(\mathbf{x}_{t_{i+1}}\mid \mathbf{x}_{0},\mathbf{x}_{1})}\left\{ d\left[ \mathbf{c}_{\theta}(\mathbf{x}_{t_{i+1}},t_{i+1}), \mathbf{c}_{\theta^{-}} (\mathbf{x}_{t_{i+1}}, t_{i+1})\right. \right.\notag\\
        &\quad-\partial_{1}\mathbf{c}_{\theta^{-}} (\mathbf{x}_{t_{i+1}},t_{i+1})(t_{i+1}-t_{i})\mathbf{u}_{t_{i+1}}(\mathbf{x}_{t_{i+1}}\mid \mathbf{x}_{0},\mathbf{x}_{1}) \notag\\
        &\quad\quad\left.\left.-\partial_{2}\mathbf{c}_{\theta^{-}} (\mathbf{x}_{t_{i+1}},t_{i+1})(t_{i+1}-t_{i}) +o(\Delta t)\right]  \right\}\notag\\
        &=\mathbb{E}_{i, p(\mathbf{x}_{0},\mathbf{x}_{1}),p(\mathbf{x}_{t_{i+1}}\mid \mathbf{x}_{0},\mathbf{x}_{1})}\left\{ d\left[ \mathbf{c}_{\theta}\left(\mathbf{x}_{t_{i+1}},t_{i+1}), \mathbf{c}_{\theta^{-}} (\mathbf{x}_{t_{i+1}}-(t_{i+1}-t_{i})\mathbf{u}_{t_{i+1}}(\mathbf{x}_{t_{i+1}}\mid \mathbf{x}_{0},\mathbf{x}_{1}), t_i \right) \right]  \right\} \notag\\
        &\quad+o(\Delta t)\notag\\
        &=\mathcal{L}_{\mathrm{GCT}}(\theta)+o(\Delta t)
    \end{align}
\endgroup
where (i) is due to the law of total expectation.

\textbf{Remark 3:} Generalized consistency training enables us to extend the application of consistency models to any forward diffusion processes or conditional ODE including those that introduce non-Gaussian noise. For example, in the Dirichlet Diffusion Score Model \citep{avdeyev_dirichlet_2023}, the forward diffusion process is a multivariate Jacobi process which transforms one-hot encoding of discrete data (e.g., DNA sequences) into Dirichlet stationary distribution. Such diffusion processes are not supported by the original consistency training approach but are feasible with generalized consistency training. We leave further applications of generalized consistency training for future work.

\textbf{Lemma 2}\label{appendix:lemma2}:
Assuming the consistency function $\mathbf{c}_{\theta}(\mathbf{x},t)$ is twice differentiable and $\partial\mathbf{c}_{\theta}(\mathbf{x},t)/\partial \mathbf{x}$ is almost everywhere nonzero, when the inverse consistency loss $\mathcal{L}_{\mathrm{ICM}}=0$, there exists a corresponding ODE vector field $\mathbf{v}_{t}^{\theta}(\mathbf{x})$ that minimized the inverse flow matching loss $\mathcal{L}_{\mathrm{IFM}}$ to $0$.

\textit{Proof:}

When the inverse consistency function is minimized to 0, we have

\begin{equation}
    \begin{aligned}
        0=\mathcal{L}_{ICM}(\theta)= \mathbb{E}_{i, p(\mathbf{x}_1), p\left(\mathbf{x}_{t_{i+1}} \mid \mathbf{x}_0=\mathbf{c}_{\theta} (\mathbf{x}_1,1)\right)} \left\|\left( \mathbf{c}_{\theta}(\mathbf{x}_{t_{i+1}},t_{i+1})-\text{stopgrad}\left(\mathbf{c}_{\theta} (\mathbf{x}_{t_{i}},t_{i}) \right) \right) \right\|
        \\\mathbf{x}_{t_{i+1}}-\mathbf{x}_{t_{i}} = \mathbf{u}_{t_{i+1}}(\mathbf{x}_{t_{i+1}}\mid \mathbf{x}_0)(t_{i+1}-t_i)
    \end{aligned}
\end{equation}
which is equivalent to
\begin{equation}
    \begin{aligned}
        0&=\mathbf{c}_{\theta}(\mathbf{x}_{t_{i+1}}, t_{i+1})-\mathbf{c}_{\theta}(\mathbf{x}_{t_{i}},t_{i})\\
        &=\partial_{1}\mathbf{c}_{\theta}(\mathbf{x}_{t_{i+1}},t_{i+1})(\mathbf{x}_{t_{i+1}}-\mathbf{x}_{t_{i}})+\partial_{2}\mathbf{c}_{\theta}(\mathbf{x}_{t_{i+1}},t_{i+1})(t_{i+1}-t_{i})+\mathbf{c}_{\theta}(\mathbf{x}_{t_{i+1}},t_{i+1})-\mathbf{c}_{\theta}(\mathbf{x}_{t_{i+1}},t_{i+1})\\
        &=\partial_{1}\mathbf{c}_{\theta}(\mathbf{x}_{t_{i+1}},t_{i+1})\mathbf{u}_{t_{i+1}}(\mathbf{x}_{t_{i+1}}\mid \mathbf{x}_0)(t_{i+1}-t_i)+\partial_{2}\mathbf{c}_{\theta}(\mathbf{x}_{t_{i+1}},t_{i+1})(t_{i+1}-t_{i}),
    \end{aligned}
\end{equation}
Then we have the connection between the learned consistency function $\mathbf{c}_{\theta}(\mathbf{x},t)$ and the conditional ODE vector field $\mathbf{u}_{t}(\mathbf{x}\mid \mathbf{x}_{0})$ where $\mathbf{c}_{\theta}(\mathbf{x}_{1},1)$ is substituted for $\mathbf{x}_{0}$:
\begin{equation}\label{eqn:relation}
    \begin{aligned}
        \partial_{1}\mathbf{c}_{\theta}(\mathbf{x},t)\mathbf{u}_{t}\left(\mathbf{x}\mid \mathbf{c}_{\theta}(\mathbf{x}_{1},1)\right)+\partial_{2}\mathbf{c}_{\theta}(\mathbf{x},t)=0
    \end{aligned}
\end{equation}

Inspired by the above result, we construct an ODE vector field as 
\begin{equation}
    \begin{aligned}
        \mathbf{v}^{\theta}_{t}(\mathbf{x})=\frac{\partial_{2}\mathbf{c}_{\theta}(\mathbf{x},t)}{\partial_{1}\mathbf{c}_{\theta}(\mathbf{x},t)}
    \end{aligned}
\end{equation}
where $\partial_{1}\mathbf{c}_{\theta}=\partial\mathbf{c}_{\theta}/\partial\mathbf{x}\neq0$ almost everywhere and for all 
$(\mathbf{x},t)$ where $\partial_{1}\mathbf{c}_{\theta}(\mathbf{x},t)=0$, we define $\mathbf{v}^{\theta}_{t}(\mathbf{x})=0$.

Now we show that $\mathbf{v}^{\theta}_{t}(\mathbf{x})$ minimized the inverse flow matching loss $\mathcal{L}_{\mathrm{IFM}}$ to 0.

\begin{equation}
    \begin{aligned}
    \mathcal{L}_{\mathrm{IFM}}(\theta)=\mathbb{E}_{t, p(\mathbf{x}_1), p\left(\mathbf{x}_{t} \mid \mathbf{x}_0=\textrm{ODE}^{\mathbf{v}^{\theta}}_{1\rightarrow 0}(\mathbf{x}_1)\right)}\left\|\mathbf{v}^{\theta}_t(\mathbf{x}_{t})-\mathbf{u}_t\left(\mathbf{x}_{t} \mid \textrm{ODE}^{\mathbf{v}^{\theta}}_{1\rightarrow 0}(\mathbf{x}_1)\right)\right\|
        \end{aligned}
\end{equation}

Firstly, we argue that $\textrm{ODE}^{\mathbf{v}^{\theta}}_{1\rightarrow 0}(\mathbf{x}_1)=\mathbf{c}_{\theta}(\mathbf{x}_{1},1)$, which can be proven by noting that the consistency function $\mathbf{c}_{\theta}(\mathbf{x},t)$ maps every point along the ODE trajectory to the same point. Consider an $N$-step ODE and two consecutive points along the trajectory, say $\mathbf{x}_{t_{i}}=\textrm{ODE}^{\mathbf{v}^{\theta}}_{1\rightarrow t_{i}}(\mathbf{x}_{1})$ and $\mathbf{x}_{t_{i+1}}=\textrm{ODE}^{\mathbf{v}^{\theta}}_{1\rightarrow t_{i+1}}(\mathbf{x}_{1})$. We have 
\begin{equation}
    \begin{aligned}
        \mathbf{c}_{\theta}(\mathbf{x}_{t_{i}},t_{i})&=\mathbf{c}_{\theta}(\mathbf{x}_{t_{i+1}}-\mathbf{v}^{\theta}_{t_{i+1}}(\mathbf{x}_{t_{i+1}})(t_{i+1}-t_i),t_{i})\\
        &=\mathbf{c}_{\theta}(\mathbf{x}_{t_{i+1}},t_{i+1})\\
        &\qquad-\partial_{1}\mathbf{c}_{\theta}(\mathbf{x}_{t_{i+1}}, t_{i+1})\mathbf{v}^{\theta}_{t_{i+1}}(\mathbf{x}_{t_{i+1}})(t_{i+1}-t_i)-\partial_{2}\mathbf{c}_{\theta}(\mathbf{x}_{t_{i+1}}, t_{i+1})(t_{i+1}-t_i)\\
        &=\mathbf{c}_{\theta}(\mathbf{x}_{t_{i+1}},t_{i+1})
    \end{aligned}
\end{equation}

where derivative terms are eliminated by the definition of $\mathbf{v}^{\theta}_{t}$. Further applying the boundary condition of the consistency function,
\begin{equation}
    \begin{aligned}
        \mathbf{c}_{\theta}(\mathbf{x}_1,1)&=\mathbf{c}_{\theta}(\mathbf{x}_{t_{N}},t_{N})\\
        &=\mathbf{x}_{t_{N}}=\mathbf{x}_{0}\\
        &=\textrm{ODE}^{\mathbf{v}^{\theta}}_{1\rightarrow 0}(\mathbf{x}_1)
    \end{aligned}
\end{equation}
where $t_N=0$.

Secondly, substituting $\mathbf{v}^{\theta}_{t}(\mathbf{x})$ into $\mathcal{L}_{\mathrm{IFM}}$, we get

\begin{equation}
    \begin{aligned}
    \mathcal{L}_{\mathrm{IFM}}(\theta)=\mathbb{E}_{t, p(\mathbf{x}_1), p\left(\mathbf{x}_{t} \mid \mathbf{x}_0=\textrm{ODE}^{\mathbf{v}^{\theta}}_{1\rightarrow 0}(\mathbf{x}_1)\right)}\left\|\frac{\partial_{2}\mathbf{c}_{\theta}(\mathbf{x}_{t},t)}{\partial_{1}\mathbf{c}_{\theta}(\mathbf{x}_{t},t)}-\mathbf{u}_t\left(\mathbf{x}_{t} \mid \textrm{ODE}^{\mathbf{v}^{\theta}}_{1\rightarrow 0}(\mathbf{x}_1)\right)\right\|
        \end{aligned}
\end{equation}

Multiplying the terms inside the norm by $\partial_{1}\mathbf{c}_{\theta}(\mathbf{x}_{t},t)$ gives
 Eq. \ref{eqn:relation}:
\begin{equation}
    \begin{aligned}
        &\qquad\partial_{2}\mathbf{c}_{\theta}(\mathbf{x}_{t},t)-\mathbf{u}_t\left(\mathbf{x}_{t} \mid \textrm{ODE}^{\mathbf{v}^{\theta}}_{1\rightarrow 0}(\mathbf{x}_1)\right)\partial_{1}\mathbf{c}_{\theta}(\mathbf{x}_{t},t)\\
        &=\partial_{2}\mathbf{c}_{\theta}(\mathbf{x}_{t},t)-\mathbf{u}_t\left(\mathbf{x}_{t} \mid \mathbf{c_{\theta}(\mathbf{x}_{1},1}\right)\partial_{1}\mathbf{c}_{\theta}(\mathbf{x}_{t},t)\\
        &=0
    \end{aligned}
\end{equation}

Thus, $\mathcal{L}_{\mathrm{IFM}}(\theta)=0$ given the constructed $\mathbf{v}^{\theta}_{t}(\mathbf{x})$.

\textbf{Remark 4:} The assumption that $\partial\mathbf{c}_{\theta}(\mathbf{x},t)/\partial\mathbf{x}\neq0$ almost everywhere guarantees that the division in the construction of $\mathbf{v}^{\theta}_{t}(\mathbf{x})$ is valid, preventing singularities except on a set of measure zero. More importantly, in neural ODEs, the ability to uniquely map states forward and backward in time is essential for defining a continuous and invertible transformation. The assumption ensures that the flow remains invertible almost everywhere, preventing singularities where trajectories might merge or become non-invertible. This property is crucial for ensuring that the learned dynamics remain well-posed.

\subsection{Introduction to denoising without ground truth}\label{appendix:denoise}
The most comparable approaches to our method are those that explicitly consider a noise distribution, including Stein’s Unbiased Risk Estimate (SURE)-based denoising methods \citep{soltanayevTrainingDeepLearning2018,metzlerUnsupervisedLearningStein2020} and Noise2Score \citep{kimNoise2ScoreTweedieApproach2021}. SURE-based denoising is applicable to independent Gaussian noise and Noise2Score is more generally applicable to exponential family noise. SURE-based denoising directly optimizes a loss motivated by SURE which provides an unbiased estimate of the true risk, which is a mean-squared error to the ground truth. Noise2Score uses Tweedie’s formula for estimating the posterior mean of an exponential family distribution with the score of the noisy distribution. The score is estimated by an approximate score estimator using a denoising autoencoder.

Another family of approaches often referred to as Noise2X is based on the assumptions of centered (zero-mean) and independent noise. Noise2Noise \citep{lehtinenNoise2NoiseLearningImage2018} requires independent noisy observations of the same ground truth data. Noise2Self \citep{batsonNoise2SelfBlindDenoising2019} is based on the statistical independence across different dimensions of the measurement, such as the independence between different pixels. Noise2Void \citep{krull_noise2void_2019} leverages the concept of blind-spot networks, which predict the value of a pixel based solely on its surrounding context. Similarly, Noise2Same \citep{xieNoise2SameOptimizingSelfSupervised2020} employs self-supervised learning using selectively masked or perturbed regions to train the model to predict unobserved values. Both of them assume independence of noise across dimensions.

\subsection{Experimental details}\label{appendix:exp}
All experiments were conducted on a server with 36 cores, 400 GB memory, and NVIDIA Tesla V100 GPUs. All models were implemented with PyTorch 2.1 \citep{paszkePyTorchImperativeStyle2019} and trained with the AdamW \citep{loshchilovDecoupledWeightDecay2019} optimizer. Model architectures and training hyperparameters are listed in Table \ref{table:model_arch}.

\begin{table}[h!]\label{table:model_arch}
\caption{Model architectures and hyperparameters}
\centering
\tabcolsep=0.07cm
\begin{tabular}{cccccccc}
\hline
dataset          & architecture          & channels                                                                              & embed\_dim           & embed\_scale         & epochs & lr               & lr schedule           \\ \hline
Navier-Stokes    & \multirow{3}{*}{MLP}  & \multirow{3}{*}{\begin{tabular}[c]{@{}c@{}}{[}256,256,\\ 256,256{]}\end{tabular}}     & \multirow{3}{*}{256} & \multirow{3}{*}{1.0} & 2000   & $5\times10^{-4}$ & \multirow{3}{*}{None} \\
8-gaussians      &                       &                                                                                       &                      &                      & 2000   & $5\times10^{-4}$ &                       \\
Single-cell      &                       &                                                                                       &                      &                      & 1000   & $1\times10^{-4}$ &                       \\ \hline
Gaussian noise   & \multirow{4}{*}{UNet} & \multirow{4}{*}{\begin{tabular}[c]{@{}c@{}}{[}128,128,\\ 256,256,512{]}\end{tabular}} & \multirow{4}{*}{512} & \multirow{4}{*}{1.0} & 3000   & $1\times10^{-4}$ & StepLR                \\
Correlated noise &                       &                                                                                       &                      &                      & 1000   & $1\times10^{-4}$ & None                  \\
Jacobi process   &                       &                                                                                       &                      &                      & 1000   & $1\times10^{-4}$ & None                  \\
FMD              &                       &                                                                                       &                      &                      & 3000   & $1\times10^{-4}$ & StepLR                \\ \hline
\end{tabular}
\end{table}

\subsubsection{Training details}\label{appendix:training}
To train IFM or ICM, we first consider a discretized time sequence $\epsilon=t_{1}<t_{2}<\dots<t_{N}=1$, where $\epsilon$ is a small positive value close to 0. We follow \cite{karrasElucidatingDesignSpace2022b} to determine the time sequence with the formula $t_{i}=\left(\epsilon^{1/\rho}+\frac{i-1}{N-1}(T^{1/\rho}-\epsilon^{1/\rho})\right)^{\rho}$, where $\rho=7$, $T=1$, and $N=11$. We choose the conditional ODE vector field as
\begin{equation}
    \begin{aligned}
        \mathbf{u}_{t_{i}}(\mathbf{x}_{t_{i}}\mid \mathbf{x}_{0},\mathbf{x}_{1})=\mathbf{x}_{1}-\mathbf{x}_{0}.
    \end{aligned}
\end{equation}
Further, the gradient of the inferred noise-free data $\mathbf{x}_{0}$ is stopped to stabilize the training process, which is
\begin{equation}
    \begin{aligned}
        \mathbf{x}_{0}=\textrm{stopgrad}\left(\textrm{ODE}^{\mathbf{v}^{\theta}}_{1\rightarrow 0}(\mathbf{x}_1)\right)
    \end{aligned}
\end{equation}
for IFM and
\begin{equation}
    \begin{aligned}
        \mathbf{x}_{0}=\textrm{stopgrad}\left(\mathbf{c}_{\theta}(\mathbf{x}_{1},1)\right)
    \end{aligned}
\end{equation}
for ICM. For ICM, the loss is weighted by
\begin{equation}
    \begin{aligned}
        \lambda(i)=t_{i+1}-t_{i}
    \end{aligned}
\end{equation}
in the same way as \cite{songImprovedTechniquesTraining2023}.

\subsubsection{Synthetic datasets}\label{appendix:syn_datasets}
\begin{figure}[h]
    \centering
    \includegraphics[width=0.8\textwidth]{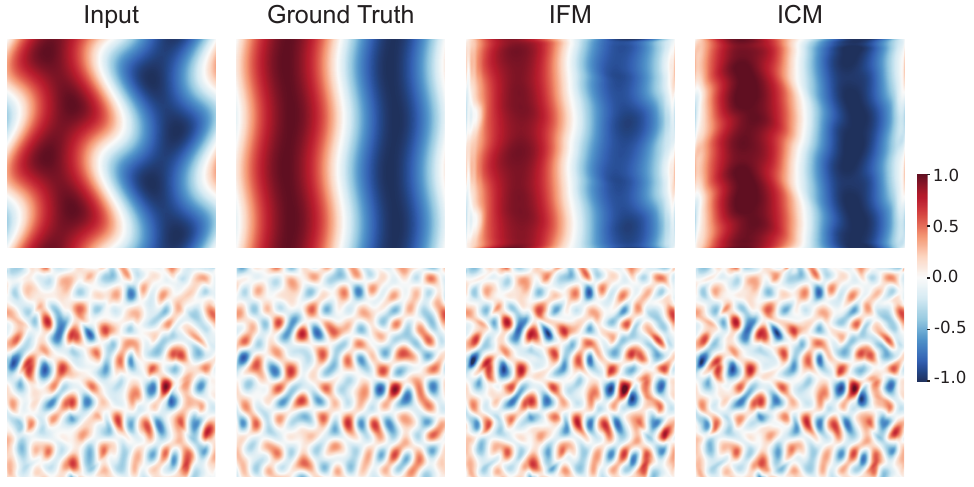}
        \vskip -0.1in
    \caption{Original Prediction of inverting Navier-Stokes fluid dynamics simulation, color indicating horizontal velocity.}
    \label{fig:suppToy}
\end{figure}

We adopted a simple form of Navier-Stokes equations which only includes the viscosity term in the fluid mechanics
\begin{equation}
    \begin{aligned}
        \rho(\frac{\partial v}{\partial t}+v\cdot \nabla v)&=-\nabla p + \mu\nabla^2 v \\
        \nabla\cdot v&=0
    \end{aligned}
\end{equation}
where $\rho$ is the density of the fluid, $v$ is the velocity, $p$ is the pressure and $\mu$ is the viscosity coefficient. For inverting the Navier-Stokes simulations, we simulated the fluid data within a 2D boundary of $[0,1]\times[0,1]$ domain from $t=0$ to $t=0.1$ with the spectral method \citep{spalart_spectral_1991}. For the upper simple case shown in Figure \ref{fig:toy}, the initial flow vector field was chosen as:
\begin{equation}
    \begin{aligned}
        \mathbf{v}_{x}&=-\sin(2\pi y)\\
        \mathbf{v}_{y}&=\sin(4\pi x)
    \end{aligned}
\end{equation}
For the bottom complex case, the initial flow vector was constructed by creating a random stream function:
\begin{equation}
    \begin{aligned}
        \psi(x,y)=\sum_{i=1}^{N}A_{i}\sin(k_{x}^{i}x)*\cos(k_{y}^{i}y)
    \end{aligned}
\end{equation}
where we choose $N=20$, $A_{i}\sim\mathcal{U}[0,2]$, $k_{x}^{i}\sim\mathcal{U}[0,10]$, and $k_{u}^{i}\sim\mathcal{U}[0,10]$. Then the flow vector field was defined as
\begin{equation}
    \begin{aligned}
        \mathbf{v}_{x}&=\frac{\partial\psi}{\partial y}\\
        \mathbf{v}_{y}&=-\frac{\partial\psi}{\partial x}.
    \end{aligned}
\end{equation}
We show the original prediction of flow fields in Figure \ref{fig:suppToy}.

The 8-gaussians is generated by adding independent gaussian noise ($\sigma=0.15$) to 8 points whose coordinates are $(0,1),(0.-1),(1,0),(-1,0),(\frac{\sqrt{2}}{2},\frac{\sqrt{2}}{2}),(\frac{\sqrt{2}}{2},-\frac{\sqrt{2}}{2}),(-\frac{\sqrt{2}}{2},\frac{\sqrt{2}}{2}),(-\frac{\sqrt{2}}{2},-\frac{\sqrt{2}}{2})$. The dataset is composed of 8000 points for training and 1600 points for testing.

We used a simple MLP-based model architecture with Gaussian Fourier time embedding in Table \ref{table:model_arch}. All methods were trained with a learning rate of $5\times10^{-4}$ for 2000 epochs. The model training took about 10 minutes.

\subsubsection{Real-world datasets}\label{appendix:realdata}

All models were trained using the BSDS500 training set with 200 images randomly cropped to the size of $256\times256$ and evaluated on the BSDS500 test set, Kodak, and Set12 with images cropped to the same size at the center. We used the same UNet-based model architecture as \cite{lehtinenNoise2NoiseLearningImage2018} with additional Gaussian Fourier time embedding listed in Table \ref{table:model_arch}.

The URL for each dataset is given:

BSDS500 \citep{arbelaezContourDetectionHierarchical2011}: \url{https://www2.eecs.berkeley.edu/Research/Projects/CS/vision/bsds/}

Kodak: \url{https://r0k.us/graphics/kodak/}

Set12 \citep{zhangGaussianDenoiserResidual2017}: \url{https://github.com/cszn/DnCNN/tree/master/testsets/Set12}

\textbf{Gaussian noise} is applied with
\begin{equation}
    \begin{aligned}
        \mathbf{x}_{1}=\mathbf{x}_{0}+\eta
    \end{aligned}
\end{equation}
where $\mathbf{x}_{0}$ is the noise-free data, $\mathbf{x}_{1}$ is a noisy observation, and $\eta\sim \mathcal{N}(0,\sigma^{2}I)$. We chose $\sigma=25$ in the experiments. All models were trained with the following setting. The total epoch was set to 3000. The learning rate was initialized to $1\times10^{-4}$ for the first 1500 epochs and was decayed to $5\times10^{-5}$ for the last 1500 epochs. The model training took about 1.5 hours.

\textbf{Correlated noise} is applied similarly to independent Gaussian noise. We adopt the method from \cite{makinenCollaborativeFilteringCorrelated2020} with
\begin{equation}
    \begin{aligned}
        \eta=\nu \circledast g
    \end{aligned}
\end{equation}
where $\nu\sim \mathcal{N}(0, \sigma^{2}I)$ and $g$ is a convolution kernel. We consider $g$ in the form of
\begin{equation}
    \begin{aligned}
        g=\frac{1}{2\pi a^{2}}\cos{|r|}\exp{(-\frac{r^{2}}{2a^{2}})}
    \end{aligned}
\end{equation}
in polar coordinates and $a$ determines the level of correlation. We generated the correlated noisy observation with $\sigma=25$ and $a=2$. All models were trained with a learning rate of $1\times10^{-4}$ for 1000 epochs. The model training took about 30 minutes.

\textbf{Jacobi process} takes the following form
\begin{equation}
    \begin{aligned}
        \mathrm{d}\mathbf{x} = \frac{s}{2}[a(1-\mathbf{x})-b\mathbf{x}]\mathrm{d}t + \sqrt{s\mathbf{x}(1-\mathbf{x})} \mathrm{d}\mathbf{w},
    \end{aligned}
\end{equation}
where $0\leq x\leq 1$, $s > 0$ is the speed factor, and $a > 0$, $b > 0$ determines the stationary distribution $\mathrm{Beta}(a, b)$. Note that when $x$ approaches $0$ or $1$, the diffusion coefficient converges to $0$ and the drift coefficient converges to $a$ or $-b$, keeping the diffusion within $[0,1]$. We used $s=1$ and $a=b=1$ and generated the noisy observation $\mathbf{x}_{1}$ with an Euler-Maruyama sampler to simulate the SDE from the initial value $\mathbf{x}_{0}$. All models were trained with a learning rate of $1\times10^{-4}$ for 1000 epochs. The model training took about 1.5 hours.

\subsubsection{Denoising microscopic data}\label{appendix:micro}
The Fluorescence Microscopy Denoising (FMD) dataset published by \cite{zhang_poisson-gaussian_2019} was downloaded from \url{https://github.com/yinhaoz/denoising-fluorescence}. We adopted the signal dependent noise model from \cite{liu_estimation_2013}
\begin{equation}
    \begin{aligned}
        g=f+f^{\gamma}\cdot u + w
    \end{aligned}
\end{equation}
to estimate the condition noise distribution where $g$ is the noisy pixel value, $f$ is the noise-free pixel value, $\gamma$ is the exponential parameter, and $u$ and $w$ are zero-mean random variables with variance $\sigma_{u}^{2}$ and $\sigma_{w}^{2}$, respectively. The variance of the noise model is
\begin{equation}
    \begin{aligned}
        \sigma^{2}=f^{2\gamma}\cdot \sigma_{u}^{2}+\sigma_{w}^{2}.
    \end{aligned}
\end{equation}
To estimate the parameters in the noise model, we split an image into $4\times4$ patches. We assume the variance within a patch is constant and approximate the noise-free pixel values of the patches by the mean values. The parameters in the noise model are estimated by the Maximum-Likelihood method.

We used the same UNet-based model architecture as \cite{lehtinenNoise2NoiseLearningImage2018} with additional Gaussian Fourier time embedding listed in Table \ref{table:model_arch}. The learning rate was initialized to $1\times10^{-4}$ for the first 1500 epochs and was decayed to $5\times10^{-5}$ for the last 1500 epochs.

\subsubsection{Denoising Single-cell genomics data}\label{appendix:scRNA}
The adult mouse brain dataset published by \cite{zeisel_molecular_2018} was downloaded from \url{https://www.ncbi.nlm.nih.gov/sra/SRP135960}.
The dentate gyrus neurogenesis dataset published by \cite{hochgernerConservedPropertiesDentate2018} was downloaded from \url{https://www.ncbi.nlm.nih.gov/geo/query/acc.cgi?acc=GSE104323}  and the neuron- and glia-related cells were kept for denoising. We preprocessed the datasets by the standard pipeline \citep{wolfSCANPYLargescaleSinglecell2018} and then performed principal component analysis. We further normalized the datasets by scaling the standard deviation of the first principal component to 1. After that, we denoised the datasets using the top 6 principal components with $\sigma=0.4$. We used a simple MLP-based model architecture with Gaussian Fourier time embedding in Table \ref{table:model_arch}. The model was trained with a learning rate of $1\times10^{-4}$ for 1000 epochs. The model training took about 5 minutes.

\subsection{Additional experiments}\label{appendix:addition}
We provide extensive experiments to measure how different levels of Gaussian noise, different noise level assumptions, and different combinations of noises affect performance. We adopted the same model architecture and training strategy as for FMD in Table \ref{table:model_arch}.
.
\subsubsection{Different levels of Gaussian noise}\label{appendix:diff_level}
We conducted experiments to evaluate the performance of our method under different intensities of Gaussian noise. We performed experiments from $\sigma=5$ to $\sigma=50$ and found that our method is robust over all noise levels we applied (Table \ref{table:diff_level}). 

\begin{table}[h!]\label{table:diff_level}
\caption{Denoising performance for different levels of Gaussian noise measured by PSNR in dB}
\centering
\begin{tabular}{ccccccccccc}
\hline
        & \multicolumn{2}{c}{$\sigma=5$} & \multicolumn{2}{c}{$\sigma=12.5$} & \multicolumn{2}{c}{$\sigma=25$} & \multicolumn{2}{c}{$\sigma=50$} & \multicolumn{2}{c}{$\sigma=75$} \\
        & Input       & Pred       & Input         & Pred        & Input        & Pred       & Input        & Pred       & Input        & Pred       \\ \hline
BSDS500 & 34.15       & 37.56            & 26.19         & 31.85             & 20.17        & 28.16            & 14.15        & 24.98            & 10.63        & 23.33            \\
Kodak   & 34.15       & 37.92            & 26.19         & 32.56             & 20.18        & 29.08            & 14.15        & 25.96            & 10.63        & 24.33            \\
Set12   & 34.15       & 37.87            & 26.20         & 32.78             & 20.16        & 29.19            & 14.13        & 25.78            & 10.63        & 23.86            \\ \hline
\end{tabular}
\end{table}

\subsubsection{Different combinations of noises}
 We considered additive Gaussian noise and multiplicative noise such as Gamma noise, Poisson noise, and Rayleigh noise, as well as their combinations and on a channel-correlated RGB dataset. We followed the noise distributions introduced in Noise2Score \citep{kimNoise2ScoreTweedieApproach2021,xie_unsupervised_2023}. For combinations of multiplicative noise and Gaussian noise, we added Gaussian noises with $\sigma=10$ to the individual multiplicative noise models. As shown in Table \ref{table:noise_com}, our method is robust over all noise type combinations we applied and superior to compared methods in most noise types.

\subsubsection{Different noise level assumptions}
We conducted experiments on data with $\sigma=25$ Gaussian noise, but training and denoising with different noise level assumptions from $\sigma=12.5$ to $\sigma=50$. Shown in Table \ref{table:diff_assump}, our method demonstrates stable performance within the range of $\sigma=25$ to $\sigma=35$, indicating that overestimating the noise level has minimal impact on the effectiveness of the model.

\subsubsection{Denoising small datasets}
In scientific discovery, the amount of data available is often very limited. To evaluate the performance of our method on small datasets, we conducted experiments on the electron microscopy denoising dataset \citep{mohan_deep_2021}. Since the original authors did not release the real experimental data, we used the simulated dataset they provided and added Poisson noise, which is the noise distribution in the real data according to their analysis. The dataset consists of 46 samples. The results indicate that our method is applicable to small datasets and outperforms other approaches in this scenario (Table \ref{table:EM}). While diffusion model is known as being data hungry, our method is efficient on sample size because it does not involve training a full generative model.

\begin{table}[ht]\label{table:noise_com}
\caption{Denoising performance for different noise distributions measured by PSNR in dB}
\centering
\begin{tabular}{ccccccc}
\hline
\textbf{Noise type}                                                                & \textbf{} & \textbf{Input} & \textbf{Noise2Void} & \textbf{Noise2Self} & \textbf{Noise2Score} & \textbf{Ours (ICM)} \\ \hline
\multirow{3}{*}{\begin{tabular}[c]{@{}c@{}}Poisson\\ $\zeta=0.01$\end{tabular}}    & BSDS500   & 23.78          & 28.29               & 28.52               & \textbf{30.53}       & 29.91               \\
                                                                                   & Kodak     & 23.60          & 28.76               & 29.36               & \textbf{31.10}       & 30.58               \\
                                                                                   & Set12     & 23.08          & 30.01               & 29.23               & \textbf{30.94}       & 30.68               \\ \hline
\multirow{3}{*}{\begin{tabular}[c]{@{}c@{}}Gamma\\ $k=100$\end{tabular}}           & BSDS500   & 26.75          & 29.17               & 27.43               & 31.14                & \textbf{32.48}      \\
                                                                                   & Kodak     & 26.67          & 30.26               & 28.26               & 31.67                & \textbf{32.97}      \\
                                                                                   & Set12     & 25.53          & 30.44               & 28.54               & 31.21                & \textbf{33.08}      \\ \hline
\multirow{3}{*}{\begin{tabular}[c]{@{}c@{}}Rayleigh\\ $\sigma=0.3$\end{tabular}}   & BSDS500   & 14.03          & 28.57               & 14.86               & 30.37                & \textbf{30.55}      \\
                                                                                   & Kodak     & 13.95          & 29.73               & 14.83               & 30.96                & \textbf{31.16}      \\
                                                                                   & Set12     & 12.81          & 29.98               & 13.74               & 30.89                & \textbf{31.17}      \\ \hline
\multirow{3}{*}{Poisson+Gaussian}                                                  & BSDS500   & 22.40          & 26.45               & 27.76               & 28.54                & \textbf{29.26}      \\
                                                                                   & Kodak     & 22.25          & 27.67               & 28.86               & 29.02                & \textbf{30.02}      \\
                                                                                   & Set12     & 21.88          & 27.81               & 29.23               & 29.10                & \textbf{30.03}      \\ \hline
\multirow{3}{*}{Gamma+Gaussian}                                                    & BSDS500   & 24.29          & 27.98               & 26.10               & 29.34                & \textbf{30.53}      \\
                                                                                   & Kodak     & 24.24          & 28.99               & 27.08               & 29.90                & \textbf{31.22}      \\
                                                                                   & Set12     & 23.62          & 29.53               & 26.84               & 29.69                & \textbf{31.27}      \\ \hline
\multirow{3}{*}{Rayleigh+Gaussian}                                                 & BSDS500   & 13.85          & 28.01               & 14.72               & 29.36                & \textbf{29.79}      \\
                                                                                   & Kodak     & 13.77          & 29.12               & 14.69               & 30.12                & \textbf{30.49}      \\
                                                                                   & Set12     & 12.78          & 26.81               & 13.59               & 29.82                & \textbf{30.50}      \\ \hline
\multirow{2}{*}{\begin{tabular}[c]{@{}c@{}}GaussianRGB\\ $\sigma=25$\end{tabular}} & BSDS500   & 20.17          & 29.72               & 27.33               & 28.28                & \textbf{29.99}      \\
                                                                                   & Kodak     & 20.17          & 30.65               & 28.21               & 28.66                & \textbf{30.73}      \\ \hline
\end{tabular}
\end{table}

\begin{table}[ht]\label{table:diff_assump}
\caption{Performance for different noise level assumptions}
\centering
\begin{tabular}{cccccccc}
\hline
        & $\sigma=12.5$ & $\sigma=15$ & $\sigma=20$ & $\sigma=25$    & $\sigma=30$ & $\sigma=35$ & $\sigma=50$ \\ \hline
BSDS500 & 21.59         & 22.43       & 24.78       & \textbf{28.16} & 28.09       & 27.55       & 25.71       \\
Kodak   & 21.62         & 22.49       & 25.03       & \textbf{29.08} & 28.99       & 28.43       & 26.66       \\
Set12   & 21.67         & 22.56       & 25.14       & \textbf{29.19} & 29.20       & 28.65       & 26.86       \\ \hline
\end{tabular}
\end{table}

\begin{table}[ht]\label{table:EM}
\caption{Performance on the electron microscopy denoising dataset}
\centering
\begin{tabular}{ccccc}
\hline
\textbf{} & Input & Noise2Void & Noise2Self & Ours (ICM) \\ \hline
PSNR      & 23.70          & 38.67               & 41.42               & \textbf{43.78}               \\ \hline
\end{tabular}
\end{table}

\subsection{Additional qualitative results}\label{appendix:qualitative}

We provide additional denoising results of the real-world datasets. Since there is not an explicit noise magnitude $\sigma$ in the Jacobi process, we did not apply the SURE-based method \citep{metzlerUnsupervisedLearningStein2020} to this task.

\begin{figure}[!h]
    \centering
    \includegraphics[width=1.0\textwidth]{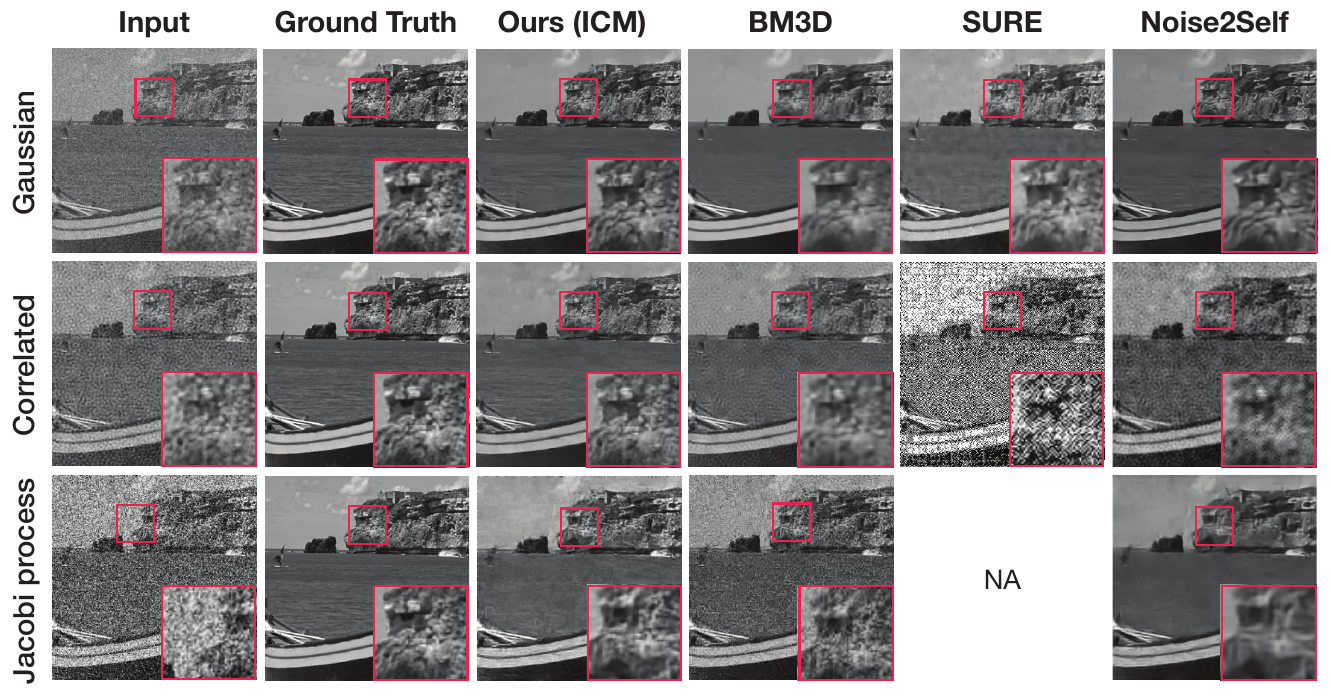}
        \vskip -0.1in
    \caption{Denoising results of BSDS500 for natural images corrupted with three types of noise distributions. Methods compared are BM3D, SURE loss, Noise2Self, and ICM.}
    \label{fig:suppNew}
\end{figure}

\begin{figure}[!h]
    \centering
    \includegraphics[width=1.0\textwidth]{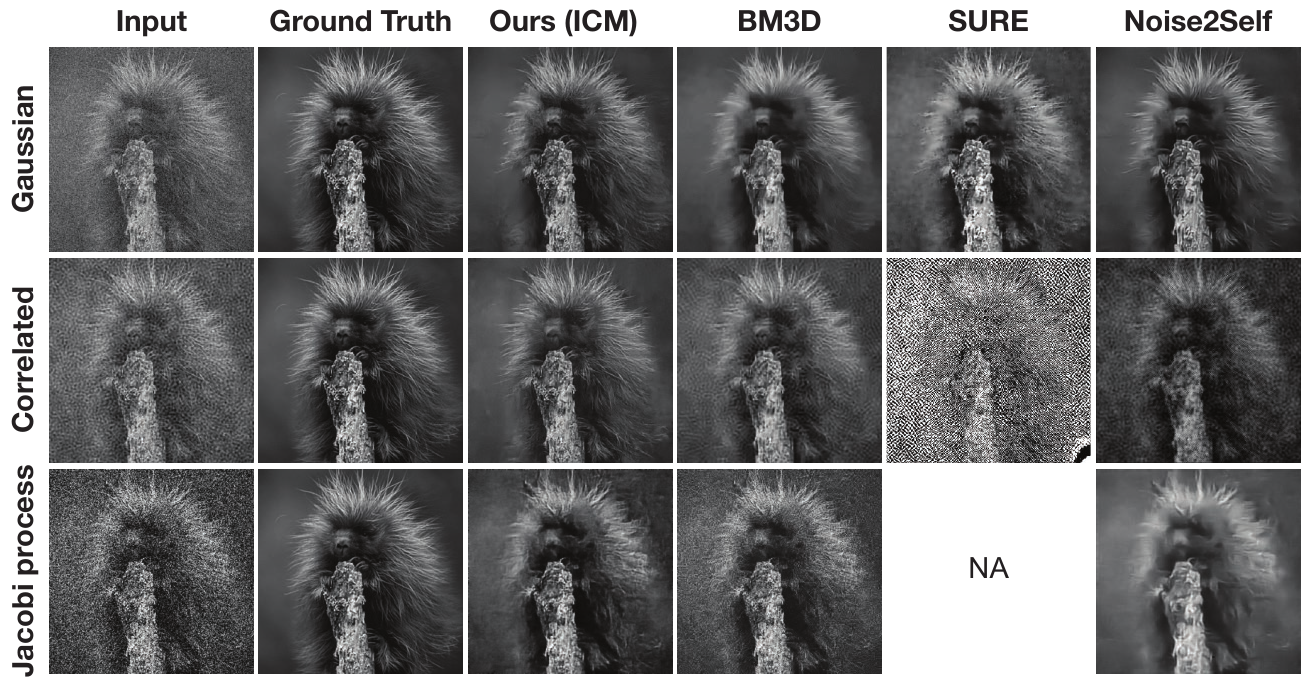}
        \vskip -0.1in
    \caption{Denoising results of BSDS500 for natural images corrupted with three types of noise distributions. Methods compared are BM3D, SURE loss, Noise2Self, and ICM.}
    \label{fig:supp86}
\end{figure}

\begin{figure}[!h]
    \centering
    \includegraphics[width=1.0\textwidth]{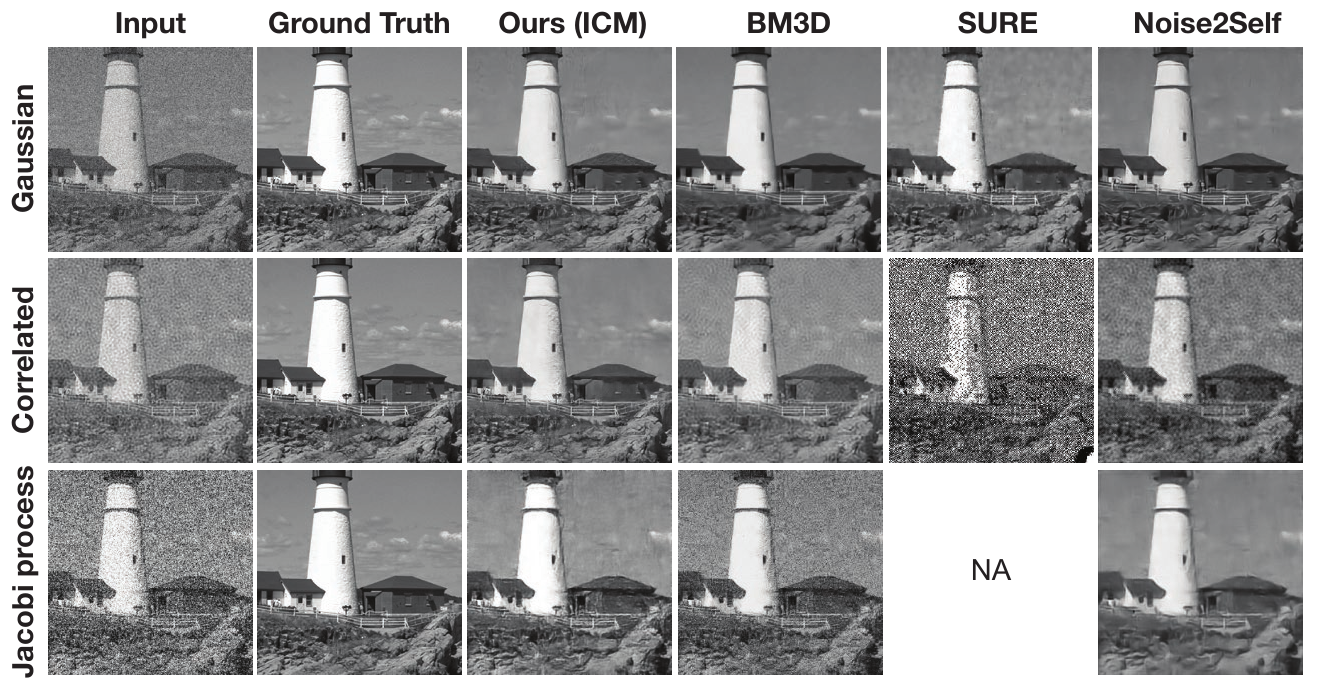}
        \vskip -0.1in
    \caption{Denoising results of Kodak for natural images corrupted with three types of noise distributions. Methods compared are BM3D, SURE loss, Noise2Self, and ICM.}
    \label{fig:suppKodak}
\end{figure}

\begin{figure}[!h]
    \centering
    \includegraphics[width=1.0\textwidth]{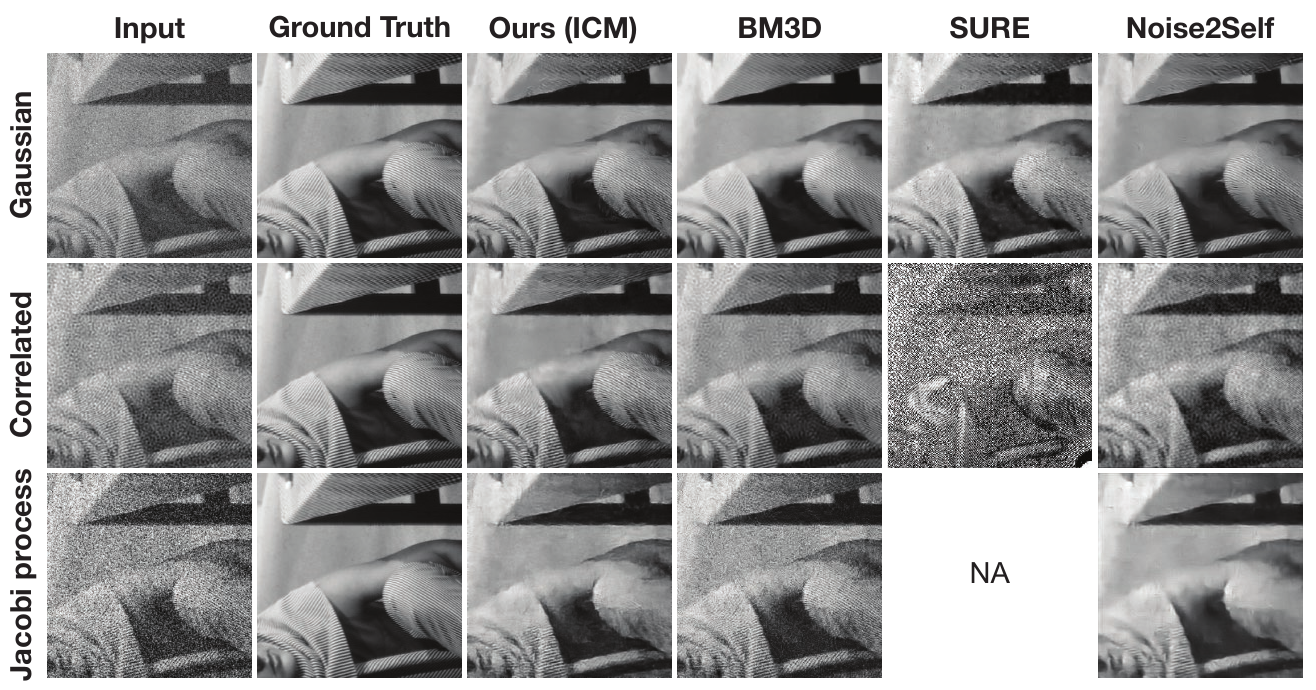}
        \vskip -0.1in
    \caption{Denoising results of Set12 for natural images corrupted with three types of noise distributions. Methods compared are BM3D, SURE loss, Noise2Self, and ICM.}
    \label{fig:suppSet12}
\end{figure}



\end{document}